\documentclass[twoside]{article}

\usepackage[accepted]{aistats2021}
\usepackage[numbers, compress]{natbib}

\usepackage{amsmath,amsfonts,bm}

\def\eqref#1{equation~\ref{#1}}

\def\1{\bm{1}}

\DeclareMathAlphabet{\mathsfit}{\encodingdefault}{\sfdefault}{m}{sl}
\SetMathAlphabet{\mathsfit}{bold}{\encodingdefault}{\sfdefault}{bx}{n}

\newcommand{\E}{\mathbb{E}}

\newcommand{\R}{\mathbb{R}}

\newcommand{\Var}{\mathrm{Var}}

\newcommand{\Cov}{\mathrm{Cov}}

\DeclareMathOperator{\Tr}{Tr}

\usepackage{booktabs}
\usepackage{enumitem}
\usepackage{array}
\usepackage{mathtools}
\usepackage{amssymb}
\usepackage{amsthm}
\usepackage{bbm}
\usepackage{bm}
\usepackage[ruled,vlined]{algorithm2e}
\usepackage{xcolor}
\usepackage{wrapfig}
\usepackage{multicol}
\usepackage{hyperref}

\newtheorem{theorem}{Theorem}

\newcommand{\inv}{^{-1}}

\begin{document}

%
\runningtitle{Shapley Value Estimation via Linear Regression}

%
\runningauthor{Covert \& Lee}

\twocolumn[

\aistatstitle{Improving KernelSHAP: Practical Shapley Value \\ Estimation via Linear Regression}

\aistatsauthor{Ian Covert \And Su-In Lee}
\aistatsaddress{University of Washington \And University of Washington} ]

\begin{abstract}%
    The Shapley value concept from cooperative game theory has become a popular technique for interpreting ML models, but efficiently estimating these values remains challenging, particularly in the model-agnostic setting. Here, we revisit the idea of estimating Shapley values via linear regression to understand and improve upon this approach. By analyzing the original KernelSHAP alongside a newly proposed unbiased version, we develop techniques to detect its convergence and calculate uncertainty estimates. We also find that the original version incurs a negligible increase in bias in exchange for significantly lower variance,
    and we propose a variance reduction technique that further accelerates the convergence of both estimators. Finally, we develop a version of KernelSHAP for stochastic cooperative games that yields fast new estimators for two global explanation methods.
\end{abstract}

\section{INTRODUCTION}

Shapley values are central to many machine learning (ML) model explanation methods (e.g., SHAP, IME, QII, Shapley Effects, Shapley Net Effects, SAGE) \cite{lundberg2017unified, lundberg2020local, vstrumbelj2014explaining, datta2016algorithmic, owen2014sobol, lipovetsky2001analysis, covert2020understanding}. Though developed in the cooperative game theory context \cite{shapley1953value}, recent work shows that Shapley values provide a powerful tool for explaining how models work when either individual features \cite{lundberg2017unified}, individual neurons in a neural network \cite{ghorbani2020neuron}, or individual samples in a dataset \cite{ghorbani2019data} are viewed as players in a cooperative game.
They have become a go-to solution for allocating credit  and quantifying contributions due to to their appealing theoretical properties.


The main challenge when using Shapley values is calculating them efficiently. A naive calculation has computational complexity that is exponential in the number of players, so numerous approaches have been proposed to accelerate their calculation.
Besides brute-force methods \cite{lipovetsky2001analysis}, other techniques include sampling-based approximations \cite{vstrumbelj2014explaining, song2016shapley, chen2018shapley, covert2020understanding}, model-specific approximations (e.g., TreeSHAP) \cite{ancona2019explaining, lundberg2020local} and a linear regression-based approximation (KernelSHAP) \cite{lundberg2017unified}.

Here, we revisit the regression-based approach to address several shortcomings in KernelSHAP. Recent work has questioned whether KernelSHAP is an unbiased estimator \cite{merrick2019explanation}, and, unlike sampling-based 
estimators \cite{castro2009polynomial, maleki2013bounding, covert2020understanding}, KernelSHAP does not provide uncertainty estimates. Furthermore, it provides no guidance on the number of samples required because its convergence properties are not well understood.

We address each of these problems, in part by building on a newly proposed unbiased version of the regression-based approach. Our contributions include:

\begin{enumerate}
    \item Deriving an unbiased version of KernelSHAP and showing empirically that the original version incurs a negligible increase in bias in exchange for significantly lower variance
    \item Showing how to detect KernelSHAP's convergence, automatically determine the number of samples required, and calculate uncertainty estimates for the results
    \item Proposing a variance reduction technique that further accelerates KernelSHAP's convergence
    \item Adapting the regression-based approach
    to stochastic cooperative games \cite{charnes1973prior} to provide fast new approximations for two global explanation methods, SAGE \cite{covert2020understanding} and Shapley Effects \cite{owen2014sobol}
\end{enumerate}

With these new insights and tools, we
offer a more practical approach to Shapley value estimation via linear regression.\footnote{\ttfamily \href{https://github.com/iancovert/shapley-regression}{https://github.com/iancovert/shapley-regression}}

\section{THE SHAPLEY VALUE}

We now provide background information on
cooperative game theory and the Shapley value.

\subsection{Cooperative Games}

A \textit{cooperative game} is a function $v: 2^d \mapsto \R$ that returns a value for each coalition (subset) $S \subseteq D$, where $D = \{1, \ldots, d\}$ represents a set of \textit{players}. 
Cooperative game theory has become increasingly important in ML because many methods frame model explanation problems in terms of cooperative games \cite{covert2020explaining}.
Notably, SHAP~\cite{lundberg2017unified}, IME \cite{vstrumbelj2014explaining} and QII \cite{datta2016algorithmic} define cooperative games that represent an individual prediction's dependence on different features.
For a model $f$ and an input $x$, SHAP (when using the marginal distribution \cite{lundberg2017unified}) analyzes the cooperative game $v_x$, defined as

\begin{equation}
    v_x(S) = \E[f(x_S, X_{D \setminus S})],
\end{equation}

where $x_S \equiv \{x_i : i \in S\}$ represents a feature subset and $X_S$ is the corresponding random variable. Two other methods, Shapley Effects \cite{owen2014sobol} and SAGE \cite{covert2020understanding}, define cooperative games that represent a model's behavior across the entire dataset. For example, given a loss function $\ell$ and response variable $Y$, SAGE uses
a cooperative game $w$ that represents the model's predictive performance given a subset of features $X_S$:

\begin{equation}
    w(S) = - \E\Big[\ell\big(\E[f(X) \; | \; X_S], Y\big)\Big]. \label{eq:sage_game}
\end{equation}

Several other techniques also frame model explanation questions in terms of cooperative games, where
a target quantity (e.g., model loss) varies as groups of players (e.g., features) are removed, and the Shapley value summarizes each player's contribution \cite{covert2020explaining}.

\subsection{Shapley Values}

The Shapley value \cite{shapley1953value}
assumes that the \textit{grand coalition} $D$ is participating and seeks to provide each player with a fair allocation of the total profit, which is represented by $v(D)$. Fair allocations must be based on each player's contribution to the profit, but a player's contribution is often difficult to define. Player $i$'s \textit{marginal contribution} to the coalition $S$ is the difference $v(S \cup \{i\}) - v(S)$, but the marginal contribution typically depends on which players $S$ are already participating.

The Shapley value resolves this problem by deriving a unique value based on a set of fairness axioms; see \cite{shapley1953value, monderer2002variations} for further detail. It can be understood as a player's average marginal contribution across all possible player orderings, and each player's Shapley value $\phi_1(v), \ldots, \phi_d(v)$ for a game $v$ is given by:

\begin{equation}
    \phi_i(v) = \frac{1}{d} \sum_{S \subseteq D \setminus \{i\}} \binom{d - 1}{|S|}\inv \Big(v(S \cup \{i\}) - v(S)\Big). \label{eq:shapley}
\end{equation}

Many ML model explanation methods can be understood in terms of ideas from cooperative game theory \cite{covert2020explaining}, but the Shapley value is especially popular and is also widely used in other fields \cite{aumann1994economic, petrosjan2003time, tarashev2016risk}.

\subsection{Weighted Least Squares Characterization}

While we can characterize the  Shapley value in many ways, the perspective most relevant to our work is viewing it as a solution to a weighted least squares problem. Many works have considered fitting simple models to cooperative games \cite{charnes1988extremal, hammer1992approximations, grabisch2000equivalent, ding2008formulas, ding2010transforms, marichal2011weighted}, particularly additive models of the form

\begin{equation*}
    u(S) = \beta_0 + \sum_{i \in S} \beta_i.
\end{equation*}

Such additive models are known as \textit{inessential games}, and although a
game $v$ may not be inessential, an inessential approximation can help summarize
each player's average contribution. Several works \cite{charnes1988extremal, hammer1992approximations, ding2008formulas} model games by solving a weighted least squares problem using a
weighting function $\mu$:

\begin{equation*}
    \min_{\beta_0, \ldots, \beta_d} \; \sum_{S \subseteq D} \mu(S) \Big(u(S) - v(S) \Big)^2.
\end{equation*}

Perhaps surprisingly, different weighting kernels $\mu$ lead to recognizable optimal regression coefficients $(\beta_1^*, \ldots \beta_d^*)$ \cite{covert2020explaining}. In particular, a carefully chosen weighting kernel yields optimal regression coefficients equal to the Shapley values \cite{charnes1988extremal, lundberg2017unified}. The Shapley kernel $\mu_{\mathrm{Sh}}$ is given by

\begin{equation*}
    \mu_{\mathrm{Sh}}(S) = \frac{d - 1}{\binom{d}{|S|}|S|(d - |S|)},
\end{equation*}

where the values $\mu_{\mathrm{Sh}}(\{\}) = \mu_{\mathrm{Sh}}(D) = \infty$ effectively enforce constraints $\beta_0 = v(\{\})$ for the intercept and $\sum_{i \in D} \beta_i = v(D) - v(\{\})$ for the sum of the coefficients. Lundberg and Lee \cite{lundberg2017unified} used this Shapley value interpretation when developing an approach to approximate SHAP values via linear regression.

\section{LINEAR REGRESSION APPROXIMATIONS}

As noted, Shapley values are difficult to calculate because they require examining each player's marginal contribution to every possible subset (Eq.~\ref{eq:shapley}). This leads to run-times  that are exponential in the number of players, so efficient approximations are of great practical importance \cite{vstrumbelj2014explaining, song2016shapley, lundberg2017unified, chen2018shapley, ancona2019explaining, lundberg2020local, covert2020understanding}. Here, we revisit the regression-based approach presented by Lundberg and Lee (KernelSHAP) \cite{lundberg2017unified} and then present an unbiased version of this approach whose properties are simpler to analyze.

\subsection{Optimization Objective}

The least squares characterization of the Shapley value suggests that we can calculate the values $\phi_1(v), \ldots, \phi_d(v)$ by solving the optimization problem

\begin{align}
    &\min_{\beta_0, \ldots, \beta_d} \; \sum_{0 < |S| < d} \mu_{\mathrm{Sh}}(S) \Big(\beta_0 + \sum_{i \in S} \beta_i - v(S) \Big)^2 \nonumber \\
    &\mathrm{s.t.} \quad \beta_0 = v(\{\}), \quad \beta_0 + \sum_{i = 1}^d \beta_i = v(D). \label{eq:constrained_problem}
\end{align}

\textbf{Notation.} We introduce new notation to make the problem easier to solve. First, we denote the non-intercept coefficients as $\beta = (\beta_1, \ldots, \beta_d) \in \R^d$. Next, we denote each subset $S \subseteq D$ using the corresponding binary vector $z \in \{0, 1\}^d$, and with tolerable abuse of notation we write $v(z) \equiv v(S)$ and $\mu_{\mathrm{Sh}}(z) \equiv \mu_{\mathrm{Sh}}(S)$ for $S = \{i : z_i = 1\}$. Lastly, we denote a distribution over $Z$ using $p(z)$, where we define $p(z) \propto \mu_{\mathrm{Sh}}(z)$ when $0 < \mathbf{1}^Tz < d$ and $p(z) = 0$ otherwise. With this, we can
rewrite the optimization problem as 

\begin{align}
    &\min_{\beta_0, \ldots, \beta_d} \; \sum_z p(z) \Big(v(\mathbf{0}) + z^T \beta - v(z) \Big)^2 \nonumber \\
    &\mathrm{s.t.}
    \quad\quad \mathbf{1}^T \beta = v(\mathbf{1}) - v(\mathbf{0}). \label{eq:problem}
\end{align}

\subsection{Dataset Sampling} \label{sec:sampling}

Solving the problem in Eq.~\ref{eq:problem} requires evaluating the cooperative game $v$ with all $2^d$ coalitions. Evaluating $v(\mathbf{0})$ and $v(\mathbf{1})$ is sufficient to ensure that the constraints are satisfied, but all values $v(z)$ for $z$ such that $0 < \mathbf{1}^Tz < d$ are required to fit the model exactly. KernelSHAP manages this challenge by subsampling a dataset and optimizing an approximate objective. We refer to this approach as \textit{dataset sampling}.
Using $n$ independent samples $z_i \sim p(Z)$ and their values $v(z_i)$, KernelSHAP solves the following
problem:

\begin{align}
    &\min_{\beta_0, \ldots, \beta_d} \; \frac{1}{n} \sum_{i = 1}^n \Big(v(\mathbf{0}) + z_i^T \beta - v(z_i) \Big)^2 \nonumber \\
    &\mathrm{s.t.}
    \quad\quad \mathbf{1}^T \beta = v(\mathbf{1}) - v(\mathbf{0}). \label{eq:approximate_problem}
\end{align}

The dataset sampling approach, also applied by LIME \cite{ribeiro2016should}, offers the flexibility to use only enough samples
to accurately approximate the objective. Given a set of samples $(z_1, \ldots, z_n)$, solving this problem is straightforward. The Lagrangian with multiplier $\nu \in \R$ is given by:

\begin{align*}
    \hat{\mathcal{L}}(\beta, \nu) = \; &\beta^T \Big(\frac{1}{n} \sum_{i = 1}^n z_iz_i^T\Big) \beta \nonumber \\
    &- 2 \beta^T \Big( \frac{1}{n} \sum_{i = 1}^n z_i \big(v(z_i) - v(\mathbf{0})\big) \Big) \nonumber \\
    &+ \frac{1}{n} \sum_{i = 1}^n \big(v(z_i) - v(\mathbf{0})\big)^2 \nonumber \\
    &+ 2\nu \big(\mathbf{1}^T \beta - v(\mathbf{1}) + v(\mathbf{0})\big).
\end{align*}

If we introduce the shorthand notation

\begin{align*}
    \hat A_n = \frac{1}{n} \sum_{i = 1}^n z_iz_i^T \quad\mathrm{and}\quad
    \hat b_n = \frac{1}{n} \sum_{i = 1}^n z_i \Big(v(z_i) - v(\mathbf{0})\Big),
\end{align*}

then we can use the problem's KKT conditions \cite{boyd2004convex} to derive the following solution:

\begin{align}
    \hat{\beta}_n &= \hat A_n\inv \Big(\hat b_n - \mathbf{1} \frac{\mathbf{1}^T \hat A_n\inv \hat b_n - v(\mathbf{1}) + v(\mathbf{0})}{\mathbf{1}^T \hat A_n\inv \mathbf{1}}\Big). \label{eq:sampled_solution}
\end{align}

This method is known as KernelSHAP \cite{lundberg2017unified}, and the implementation in the SHAP repository\footnote{\url{http://github.com/slundberg/shap}} also allows for regularization terms in the approximate objective (Eq.~\ref{eq:approximate_problem}), such as the $\ell_1$ penalty \cite{tibshirani1996regression}. While this approach is intuitive and simple to implement, the estimator $\hat{\beta}_n$ is surprisingly difficult to characterize. As we show in Section~\ref{sec:estimator_properties}, it is unclear whether it is unbiased, and understanding its variance and rate of convergence is not straightforward. Therefore, we derive an alternative approach that is simpler to analyze.

\subsection{An Exact Estimator} \label{sec:exact}

Consider the solution to the problem that uses all $2^d$ player coalitions (Eq.~\ref{eq:problem}). Rather than finding an \textit{exact solution to an approximate problem} (Section~\ref{sec:sampling}), we now derive an \textit{approximate solution to the exact problem}. The full problem's Lagrangian is given by

\begin{align*}
    \mathcal{L}(\beta, \nu) = \; &\beta^T \E[ZZ^T] \beta \nonumber \\
    &- 2 \beta^T \E\Big[Z \big(v(Z) - v(\mathbf{0})\big) \Big] \nonumber \\
    &+ \E\Big[\big(v(Z) - v(\mathbf{0})\big)^2\Big] \nonumber \\
    &+ 2\nu \big(\mathbf{1}^T \beta - v(\mathbf{1}) + v(\mathbf{0})\big),
\end{align*}

where we now consider $Z$ to be a random variable distributed according to $p(Z)$. Using the shorthand notation

\begin{align*}
    A = \E[ZZ^T] \quad\mathrm{and}\quad b = \E\Big[Z \big(v(Z) - v(\mathbf{0})\big)\Big],
\end{align*}

we can write the solution to the exact problem as:

\begin{equation}
    \beta^* = A\inv \Big(b - \mathbf{1} \frac{\mathbf{1}^T A\inv b - v(\mathbf{1}) + v(\mathbf{0})}{\mathbf{1}^T A\inv \mathbf{1}}\Big). \label{eq:solution}
\end{equation}

Due to our setup of the optimization problem, we have the property that $\beta_i^* = \phi_i(v)$. Unfortunately, we cannot evaluate this expression in practice without evaluating $v$ for all $2^d$ coalitions $S \subseteq D$.

However, knowledge of $p(Z)$ means that $A \in \R^{d \times d}$ can be calculated exactly and efficiently. To see this, note that $(ZZ^T)_{ij} = Z_iZ_j = \mathbbm{1}(Z_i = Z_j = 1)$. Therefore, to calculate $A$, we need to estimate only  $p(Z_i = 1)$ for diagonal values $A_{ii}$ and $p(Z_i = Z_j = 1)$ for off-diagonal values $A_{ij}$. See Appendix~\ref{app:exact} for their derivations.

Since $b$ cannot be calculated exactly and efficiently due to its dependence on $v$, this suggests that we should use $A$'s exact form and approximate $\beta^*$ by estimating (only) $b$. We propose the following estimator for $b$:

\begin{equation*}
    \bar b_n = \frac{1}{n} \sum_{i = 1}^n z_i v(z_i) - \E[Z] v(\mathbf{0}).
\end{equation*}

Using this, we arrive at an
alternative to the original KernelSHAP estimator, which we refer to as \textit{unbiased KernelSHAP}:

\begin{align}
    \bar{\beta}_n = A\inv \Big(\bar b_n - \mathbf{1} \frac{\mathbf{1}^T A\inv \bar b_n - v(\mathbf{1}) + v(\mathbf{0})}{\mathbf{1}^T A\inv \mathbf{1}}\Big). \label{eq:modified_solution}
\end{align}

In the next section, we compare these two approaches both theoretically and empirically.

\section{ESTIMATOR PROPERTIES} \label{sec:estimator_properties}

We now analyze the consistency, bias and variance properties of the Shapley value estimators, and we consider how to detect, forecast, and accelerate their convergence.

\subsection{Consistency, Bias and Variance} \label{sec:properties}

A \textit{consistent} estimator is one that converges to the correct Shapley values $\beta^*$ given a sufficiently large number of samples. If the game $v$ has bounded value, then the strong law of large numbers implies that

\begin{equation*}
    \lim_{n \to \infty} \hat A_n = A \quad\mathrm{and}\quad \lim_{n \to \infty} \hat b_n = \lim_{n \to \infty} \bar b_n = b,
\end{equation*}

where the convergence is almost sure. From this, we see that both estimators are consistent:

\begin{equation*}
    \lim_{n \to \infty} \; \hat{\beta}_n = \lim_{n \to \infty} \; \bar{\beta}_n = \beta^*.
\end{equation*}

Next, an \textit{unbiased} estimator is one whose expectation is equal to the correct Shapley values $\beta^*$. This is difficult to verify for the KernelSHAP estimator $\hat{\beta}_n$ due to the interaction between $\hat A_n$ and $\hat b_n$ (see Eq.~\ref{eq:sampled_solution}). Both $\hat A_n$ and $\hat b_n$ are unbiased, but terms such as $\E[\hat A_n\inv \hat b_n]$ and $\E[\hat A_n\inv \mathbf{1} \mathbf{1}^T \hat A_n\inv \hat b_n / (\mathbf{1}^T \hat A_n\inv \mathbf{1})]$ are difficult to characterize. To make any claims about KernelSHAP's bias, we rely instead on empirical observations.

In contrast, it is easy to see that the alternative estimator $\bar \beta_n$ is unbiased. Because of its linear dependence on $\bar b_n$ and the fact that $\E[\bar b_n] = b$, we can see that

\begin{equation*}
    \E[\bar{\beta}_n] = \beta^*.
\end{equation*}

We therefore conclude that the alternative estimator $\bar{\beta}_n$ is both consistent and unbiased, whereas the original KernelSHAP ($\hat{\beta}_n$) is only provably consistent. It is for this reason that we refer to
$\bar \beta_n$ as \textit{unbiased KernelSHAP}.

Regarding the estimators' variance, unbiased KernelSHAP is once again simpler to characterize. The values $\bar \beta_n$ are a function of $\bar b_n$, and the multivariate central limit theorem (CLT)  \cite{van2000asymptotic} asserts that $\bar b_n$ converges in distribution to a multivariate Gaussian, or

\begin{equation}
    \bar b_n \sqrt{n} \xrightarrow{D} \mathcal{N}(b, \Sigma_{\bar b}) \label{eq:clt},
\end{equation}

where $\Sigma_{\bar b} = \Cov\big(Zv(Z)\big)$. This implies that for the estimator $\bar \beta_n$, we have the convergence property

\begin{equation}
    \bar{\beta}_n \sqrt{n} \xrightarrow{D} \mathcal{N}(\beta^*, \Sigma_{\bar \beta}),
\end{equation}

where, due to its linear dependence on $\bar b_n$ (see Eq.~\ref{eq:modified_solution}), we have the covariance $\Sigma_{\bar \beta}$ given by

\begin{align}
    \Sigma_{\bar \beta} &= C \Sigma_{\bar b} C^T \\
    C &= A\inv - \frac{A\inv \mathbf{1} \mathbf{1}^T A\inv}{\mathbf{1}^T A\inv \mathbf{1}}. \label{eq:C}
\end{align}

This allows us to reason about unbiased KernelSHAP's asymptotic distribution. In particular, we remark that $\bar \beta_n$ has variance that reduces at a rate of $\mathcal{O}(\frac{1}{n})$.

In comparison, the original KernelSHAP estimator $\hat{\beta}_n$ is difficult to analyze due to the interaction between the $\hat A_n$ and $\hat b_n$ terms. We can apply the 
CLT to either term individually, but reasoning about $\hat{\beta}_n$'s distribution or variance remains challenging.

To facilitate our analysis of KernelSHAP, we present a simple experiment to compare the two estimators. We approximated the SHAP values for an individual prediction in the census income dataset \cite{lichman2013uci} and empirically calculated the mean squared error relative to the true SHAP values\footnote{The true SHAP values use a sufficient number of samples to ensure convergence (see Section~\ref{sec:convergence}).} across 250 runs. We then decomposed the error into bias and variance terms as follows:

\begin{equation*}
    \underbrace{\E\big[||\hat \beta_n - \beta^*||^2\big]}_{\text{Error}} = \underbrace{\E\big[||\hat \beta_n - \E[\hat \beta_n]||^2\big]}_{\text{Variance}} + \underbrace{\big|\big|\E[\hat \beta_n] - \beta^*\big|\big|^2.}_{\text{Bias}}
\end{equation*}

\begin{figure} 
\begin{center}
\includegraphics[width=\columnwidth]{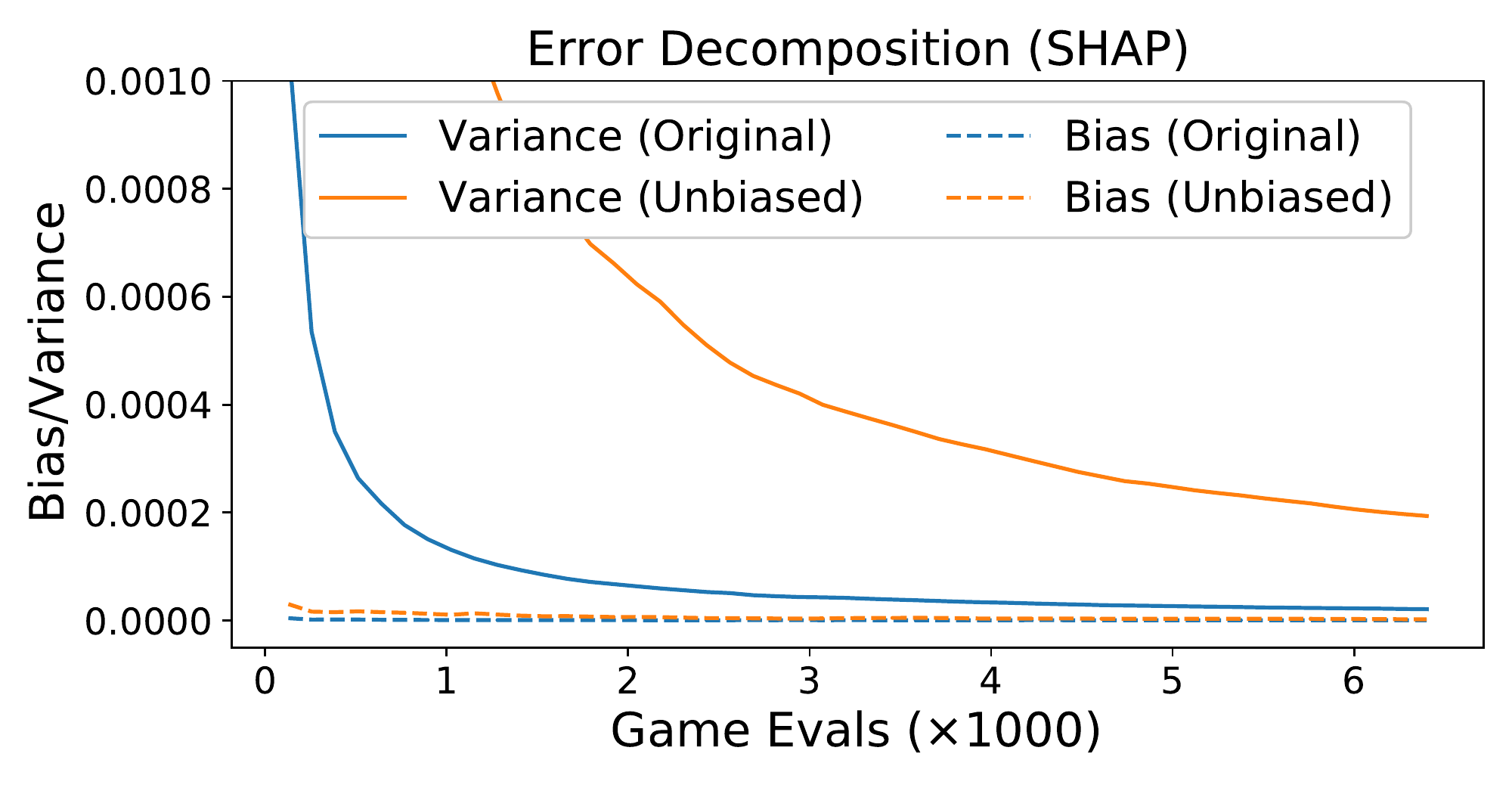}
\vskip -0.2in
\caption{SHAP error decomposition for the original and unbiased KernelSHAP estimators. The bias and variance are calculated empirically across 250 runs.}
\label{fig:error}
\end{center}
\vskip -0.1in
\end{figure}

Figure~\ref{fig:error} shows that the error for both estimators is dominated by variance rather than bias.\footnote{The unbiased approach appears to have higher bias due to estimation error, but its bias is provably zero.} It also shows that KernelSHAP incurs virtually no bias in exchange for significantly lower variance. In Appendix~\ref{app:convergence_experiments}, we provide global measures of the bias and variance to confirm these observations across multiple examples and two other datasets. This suggests that although KernelSHAP is more difficult to analyze theoretically, it should be used in practice because its bias is negligible
and it converges faster.

\subsection{Variance Reduction via Paired Sampling} \label{sec:var_reduction}

Having
analyzed each estimator's properties, we now consider whether their convergence can be accelerated. We propose a simple variance reduction technique that leads to significantly faster convergence in practice.

When sampling $n$ subsets according to the distribution $z_i \sim p(Z)$, we suggest a \textit{paired sampling} strategy where each sample $z_i$ is paired with its complement\footnote{We call $\mathbf{1} - z$ the \textit{complement} because it is the binary vector for $D \setminus S$, where $S$ corresponds to $z$.} $\mathbf{1} - z_i$. To show why this approach accelerates convergence, we focus on unbiased KernelSHAP, which is easier to analyze theoretically.

When estimating $b$ for unbiased KernelSHAP ($\bar \beta_n$), consider using the following modified estimator that combines $z_i$ with $\mathbf{1} - z_i$:

\vskip - 0.1in

\begin{equation}
    \check b_{n} = \frac{1}{2n} \sum_{i = 1}^n \big( z_i v(z_i) + (\mathbf{1} {-} z_i) v(\mathbf{1} {-} z_i) - v(\mathbf{0}) \big).
\end{equation}

Substituting this into unbiased KernelSHAP (Eq.~\ref{eq:modified_solution}) yields a new estimator $\check{\beta}_n$ that preserves the properties of being both consistent and unbiased:

\vskip - 0.1in

\begin{equation}
    \check{\beta}_{n} = A\inv \Big(\check b_{n} - \mathbf{1} \frac{\mathbf{1}^T A\inv \check b_{n} - v(\mathbf{1}) + v(\mathbf{0})}{\mathbf{1}^T A\inv \mathbf{1}}\Big). \label{eq:reduced_solution}
\end{equation}

For games $v$ that satisfy a specific condition, we can guarantee that this sampling approach leads to $\check \beta_n$ having lower variance than $\bar \beta_n$, even when we account for $\check b_n$ requiring twice as many cooperative game evaluations as $\bar b_n$ (see proof in Appendix~\ref{app:variance_reduction}).

\vskip 0.1in

\begin{theorem} \label{thm:confidence}
    The difference between the covariance matrices for the estimators $\bar \beta_{2n}$ and $\check \beta_n$ is given by
    
    \vskip - 0.05in
    
    \begin{equation*}
        \Cov(\bar \beta_{2n}) - \Cov(\check \beta_n) = \frac{1}{2n} C G_v C^T,
    \end{equation*}
    
    \vskip - 0.05in
    
    where $G_v$ is a property of the game $v$, defined as
    
    \vskip - 0.05in
    
    \begin{equation*}
        G_v = - \Cov\Big(Zv(Z), (\mathbf{1} - Z) v(\mathbf{1} - Z)\Big).
    \end{equation*}
    
    \vskip - 0.05in
    
    For sufficiently large $n$, $G_v \succeq 0$ guarantees that the Gaussian confidence ellipsoid $\bar E_{2n, \alpha}$ for $\bar \beta_{2n}$ contains the corresponding confidence ellipsoid $\check E_{n, \alpha}$ for $\check \beta_{n}$, or $\check E_{n, \alpha} \subseteq \bar E_{2n, \alpha}$,
    at any confidence level $\alpha \in (0, 1)$.
\end{theorem}

Theorem~\ref{thm:confidence} shows that $\check \beta_n$ is a more precise estimator than $\bar \beta_{2n}$ when the condition $G_v \succeq 0$ is satisfied (i.e., $G_v$ is positive semi-definite). This may not hold in the general case, but in Appendix~\ref{app:variance_reduction} we show that a weaker condition holds for \textit{all games}: the diagonal values of $G_v$ satisfy $(G_v)_{ii} \geq 0$ for any game $v$. Geometrically, this weaker condition means that $\bar E_{2n, \alpha}$ extends beyond $\check E_{n, \alpha}$ in the axis-aligned directions.

\begin{figure} 
\begin{center}
\includegraphics[width=\columnwidth]{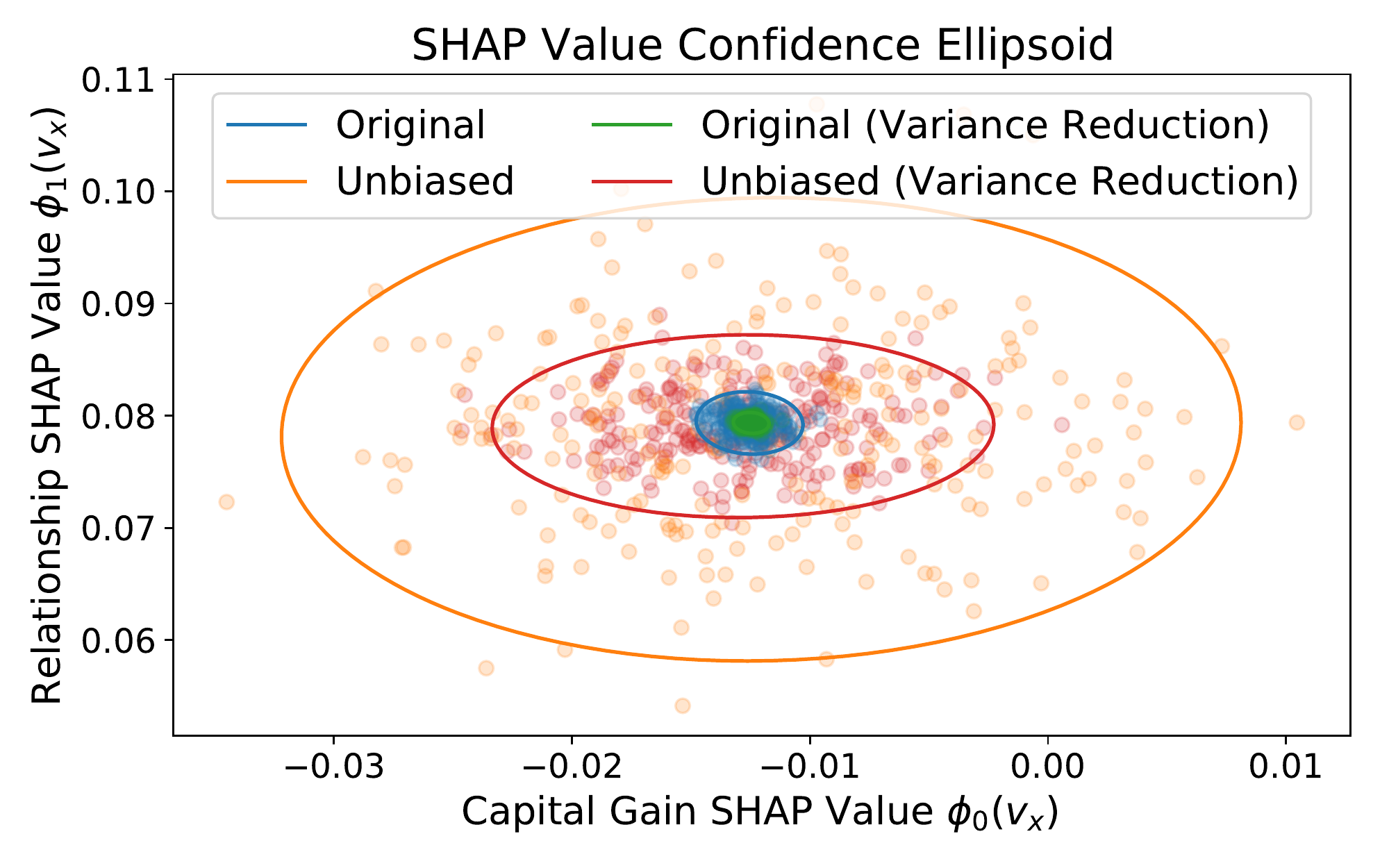}
\vskip -0.2in
\caption{Gaussian 95\% confidence ellipsoids for two SHAP values in a census income prediction (from 250 runs).
The estimators use
an equal
number of samples.}
\label{fig:confidence}
\end{center}
\vskip -0.2in
\end{figure}

Figure~\ref{fig:confidence} illustrates the result of Theorem~\ref{thm:confidence} by showing empirical 95\% confidence ellipsoids for two SHAP values. Although a comparable condition is difficult to derive for the original KernelSHAP estimator ($\hat \beta_n$), we find that the paired sampling approach yields a similar reduction in variance. Our experiments provide further evidence that this approach accelerates convergence for both estimators (Section~\ref{sec:experiments}).

\subsection{Convergence Detection and Forecasting} \label{sec:convergence}

One of KernelSHAP's practical shortcomings is its lack of guidance on the number of samples required to obtain accurate estimates. We address this problem by developing an approach for convergence detection and forecasting.

Previously, we showed that unbiased KernelSHAP ($\bar \beta_n$) has variance that reduces at a rate $\mathcal{O}(\frac{1}{n})$ (Eq.~\ref{eq:clt}). Furthermore, its variance is simple to estimate in practice: we require only an empirical estimate $\hat \Sigma_{\bar b}$ of $\Sigma_{\bar b}$ (defined above), which we can calculate using an online algorithm, such as Welford's \cite{welford1962note}.

We also showed that the original KernelSHAP ($\hat \beta_n$) is difficult to characterize, but its variance is empirically lower than the unbiased version. Understanding its variance is useful for convergence detection, so we propose an approach for approximating it. Based on the results in Figure~\ref{fig:error}, we may hypothesize that KernelSHAP's variance reduces at the same rate of $\mathcal{O}(\frac{1}{n})$; in Appendix~\ref{app:convergence_experiments}, we examine this by plotting the product of the variance and the number of samples over the course of estimation. We find that the product is \textit{constant} as the sample number increases, which suggests that the $\mathcal{O}(\frac{1}{n})$ rate holds in practice. This property is difficult to prove formally, but it can be used for simple variance approximation.

When running KernelSHAP, we suggest estimating the variance by selecting an intermediate value $m$ such that $m << n$ and calculating multiple independent estimates $\hat \beta_m$ while accumulating samples for $\hat \beta_n$. For any $n$, we can then approximate $\Cov(\hat \beta_n)$ as

\begin{equation*}
    \Cov(\hat \beta_n) \approx \frac{m}{n}\Cov(\hat \beta_m),
\end{equation*}

where $\Cov(\hat \beta_m)$ is estimated empirically using the multiple independent estimates $\hat \beta_m$. This online approach has a negligible impact on the algorithm's run-time, and the covariance estimate can be used to provide confidence intervals for the final results.

Whether we use the original or unbiased version of KernelSHAP, the estimator's covariance at a given value of $n$ lets us both detect and forecast convergence. For detection, we propose stopping at the current value $n$ when the largest standard deviation is a sufficiently small portion $t$ (e.g., $t = 0.01$) of the gap between the largest and smallest Shapley value estimates. For unbiased KernelSHAP, this criterion is equivalent to:

\begin{equation*}
    \max_i \sqrt{\frac{1}{n}(\hat \Sigma_{\bar \beta})_{ii}} < t \Big( \max_i \; (\bar{\beta}_n)_i - \min_i \; (\bar{\beta}_n)_i \Big).
\end{equation*}

To forecast the number of samples required to reach convergence, we again invoke
the property that estimates have variance that reduces at a rate of $\mathcal{O}(\frac{1}{n})$. Given a value $t$ and estimates $\bar{\beta}_n$ and $\hat \Sigma_{\bar \beta}$, the approximate number of samples $\hat N$ required is:

\begin{equation*}
    \hat N = \frac{1}{t^2} \Big(\frac{\max_i \; \sqrt{(\hat \Sigma_{\bar \beta})_{ii}}}{\max_i \; (\bar{\beta}_n)_i - \min_i \; (\bar{\beta}_n)_i}\Big)^2. \label{eq:forecast}
\end{equation*}

This allows us to forecast the time until convergence at any point during the algorithm. The forecast is expected to become more accurate as the estimated terms become more precise.

Our approach is theoretically grounded for unbiased KernelSHAP, and the approximate approach for the standard version of KernelSHAP relies only on the assumption that $\Cov(\hat \beta_n)$ reduces at a rate of $\mathcal{O}(\frac{1}{n})$. Appendix~\ref{app:algorithms} shows algorithms for both approaches, which illustrate both variance reduction and convergence detection techniques.

\begin{figure*} 
\begin{center}
\includegraphics[width=1.7\columnwidth]{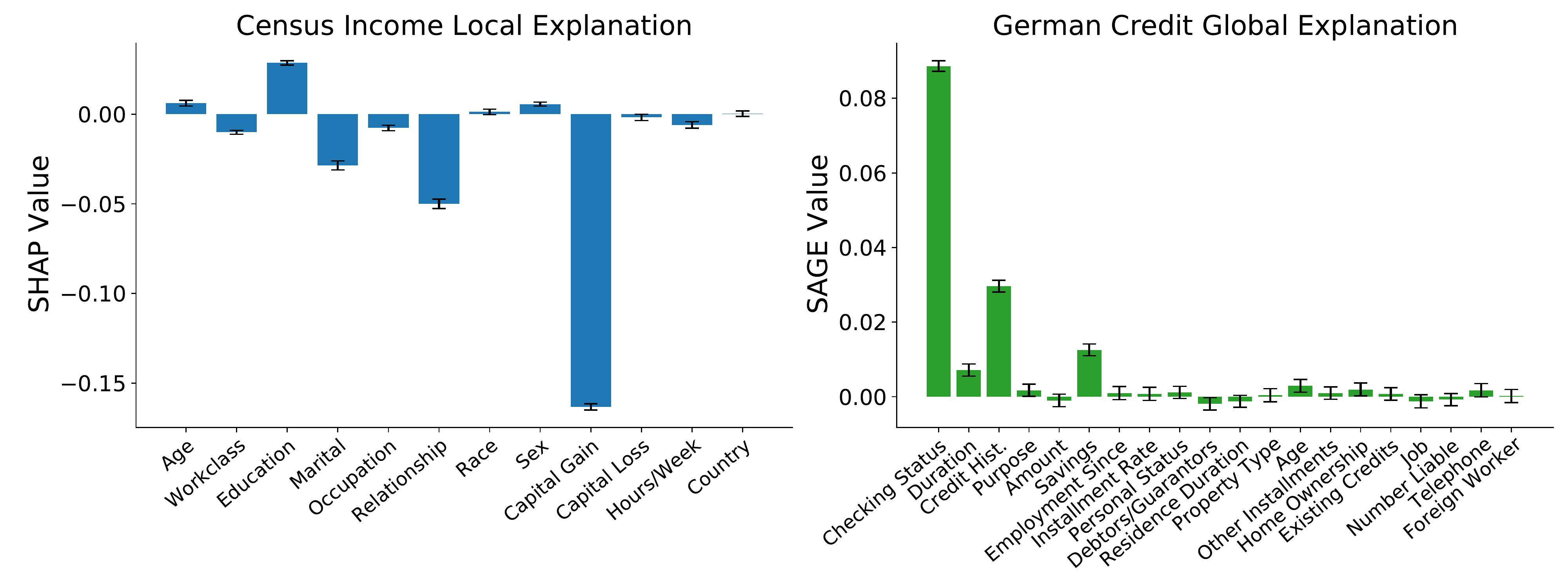}
\vskip -0.2in
\caption{Shapley value-based explanations with 95\% uncertainty estimates. Left: SHAP values for a single prediction with the census income dataset. Right: SAGE values for the German credit dataset.}
\label{fig:examples}
\end{center}
\vskip -0.1in
\end{figure*}

\section{STOCHASTIC COOPERATIVE GAMES}

We have thus far focused on developing a regression-based approach to estimate Shapley values for any cooperative game. We now discuss how to adapt this approach to stochastic cooperative games, which leads to fast estimators for two global explanation methods.

\subsection{Stochastic Cooperative Games}

\textit{Stochastic cooperative games} return a random value for each coalition of participating players $S \subseteq D$. Such games are represented by a function $V$ that maps coalitions to a \textit{distribution} of possible outcomes, so that $V(S)$ is a random variable \cite{charnes1973prior, charnes1976coalitional}. 

To aid our presentation, we assume that the uncertainty in the game can be represented by an exogenous random variable $U$. The game can then be denoted by $V(S, U)$, where $V(\cdot, U)$ is a deterministic function of $S$ for any fixed value of the variable $U$.

Stochastic cooperative games provide a useful tool for understanding two global explanation methods, SAGE \cite{covert2020understanding} and Shapley Effects \cite{owen2014sobol}. To see why, assume an exogenous variable $U = (X, Y)$ that represents a random input-label pair, and consider the following game:

\begin{equation}
    W(S, X, Y) = - \ell\Big(\E\big[f(X) | X_S\big], Y\Big).
\end{equation}

The game $W$ evaluates the (negated) loss with respect to the label $Y$ given a prediction that depends only on the features $X_S$. The cooperative game used by SAGE can be understood as the expectation of this game, or $w(S) = \E_{XY}\big[W(S, X, Y)\big]$ (see Eq.~\ref{eq:sage_game}). Shapley Effects is based on the expectation of a similar game, where the loss is evaluated with respect to the full model prediction $f(X)$ (see Appendix~\ref{app:shapley_effects}). As we show next, an approximation approach tailored to this setting yields significantly faster estimators for these methods.

\subsection{Generalizing the Shapley Value} \label{sec:generalizing}

It is natural to assign values to players in stochastic cooperative games like we do for deterministic games. We propose a simple generalization of the Shapley value for games $V(S, U)$ that averages a player's marginal contributions over both (i) player orderings and (ii) values of the exogenous variable $U$:

\begin{equation*}
    \phi_i(V) = \frac{1}{d} {\sum_{S {\subseteq} D {\setminus} \{i\}}} \binom{d {-} 1}{|S|}^{{-}1} \E_U \Big[ V(S \cup \{i\}, U) {-} V(S, U) \Big]. \label{eq:stochastic_shapley}
\end{equation*}

Due to the linearity property of Shapley values \cite{shapley1953value, monderer2002variations}, the following sets of values are equivalent:

\begin{enumerate}
    \item The Shapley values of the game's expectation $\bar v(S) = \E_U[V(S, U)]$, or $\phi_i(\bar v)$
    \item The expected Shapley values of games with fixed $U$, or $\E_U[\phi_i(v_U)]$ where $v_u(S) = V(S, u)$
    \item Our generalization of Shapley values to the stochastic cooperative game $V(S, U)$, or $\phi_i(V)$
\end{enumerate}

The first two list items suggest ways of calculating the values $\phi_i(V)$ using tools designed for deterministic cooperative games. However, the expectation $\E_U\big[V(S, U)\big]$ may be slow to evaluate (e.g., if it is across an entire dataset), and calculating Shapley values separately for each value of $U$ would be intractable if $U$ has many possible values. We therefore introduce a third approach.

\subsection{Shapley Value Approximation for Stochastic Cooperative Games} \label{sec:stochastic_kernelshap}

We now propose a fast, regression-based approach for calculating the generalized Shapley values $\phi_i(V)$ of stochastic cooperative games $V(S, U)$. Fortunately, it requires only a simple modification of the preceding approaches.

\begin{figure} 
\begin{center}
\includegraphics[trim=1.4cm 1.1cm 0.5cm 0.5cm, clip=true, width=\columnwidth]{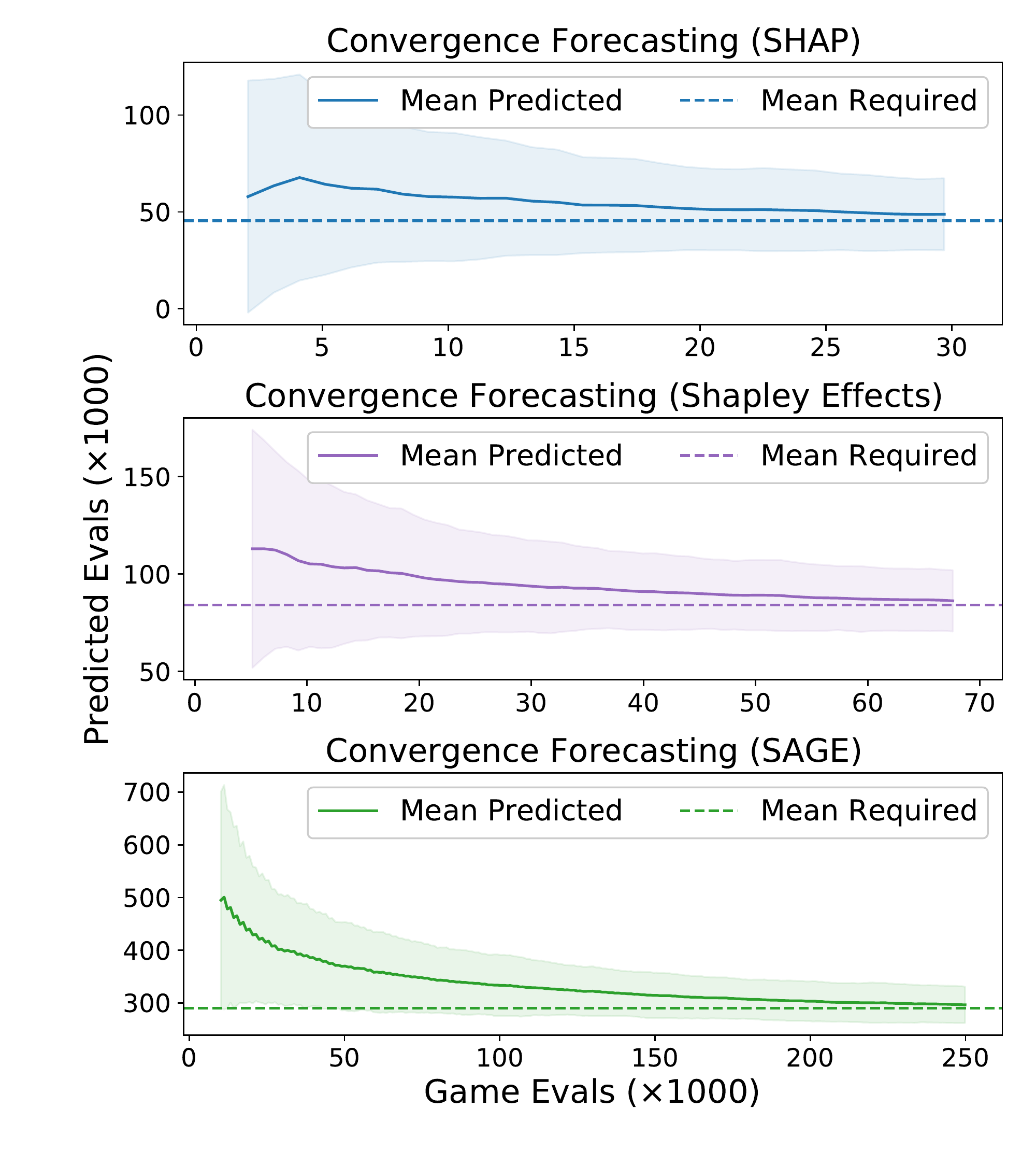}
\caption{Convergence forecasting for SHAP, Shapley Effects and SAGE. The required number of samples is compared with the predicted number across 100 runs (with 90\% confidence intervals displayed).}
\label{fig:forecasting}
\end{center}
\vskip -0.2in
\end{figure}

First, we must calculate the values $\E_U\big[ V(\mathbf{1}, U) \big]$ and $\E_U \big[ V(\mathbf{0}, U) \big]$ for the grand coalition and the empty coalition. Next, we replace our previous $b$ estimators ($\hat b_n$ and $\bar b_n$) with estimators that use $n$ pairs of independent samples $z_i \sim p(Z)$ and $u_i \sim p(U)$. To adapt the original KernelSHAP to this setting, we use

\begin{equation*}
    \tilde b_n = \frac{1}{2} \sum_{i = 1}^n z_i \big(  V(z_i, u_i) - \E_U\big[V(\mathbf{0}, U)\big] \big).
\end{equation*}

We then substitute this into the KernelSHAP estimator, as follows:

\begin{equation}
    \tilde \beta_n = \hat A_n\inv \Big(\tilde b_{n} - \mathbf{1} \frac{\mathbf{1}^T \hat A_n\inv \tilde b_{n} - v(\mathbf{1}) + v(\mathbf{0})}{\mathbf{1}^T \hat A_n\inv \mathbf{1}}\Big). \label{eq:stochastic_solution}
\end{equation}

By the same argument used in Section~\ref{sec:sampling}, this approach estimates a solution to the weighted least squares problem whose optimal solution is the generalized Shapley values $\phi_i(V)$. This adaptation of KernelSHAP is consistent, and the analogous version of unbiased KernelSHAP is consistent and unbiased (see Appendix~\ref{app:stochastic}). These can be run with our paired sampling approach, and we can also provide uncertainty estimates and detect convergence (Section~\ref{sec:estimator_properties}).

\section{EXPERIMENTS} \label{sec:experiments}

We conducted experiments with four datasets to demonstrate the advantages of our Shapley value estimation approach. We used the census income dataset \cite{lichman2013uci}, the Portuguese bank marketing dataset \cite{moro2014data}, the German credit dataset \cite{lichman2013uci}, and a breast cancer (BRCA) subtype classification dataset \cite{berger2018comprehensive}. To avoid overfitting with the BRCA data, we analyzed a random subset of 100 out of 17,814 genes (Appendix~\ref{app:experiments}). We trained a LightGBM model \cite{ke2017lightgbm} for the census data, CatBoost \cite{prokhorenkova2018catboost} for the credit and bank data, and logistic regression for the BRCA data. Code for our experiments is available online.

\begin{figure} 
\begin{center}
\includegraphics[trim=0.8cm 0cm 0.4cm 0.2cm, clip=true, width=\columnwidth]{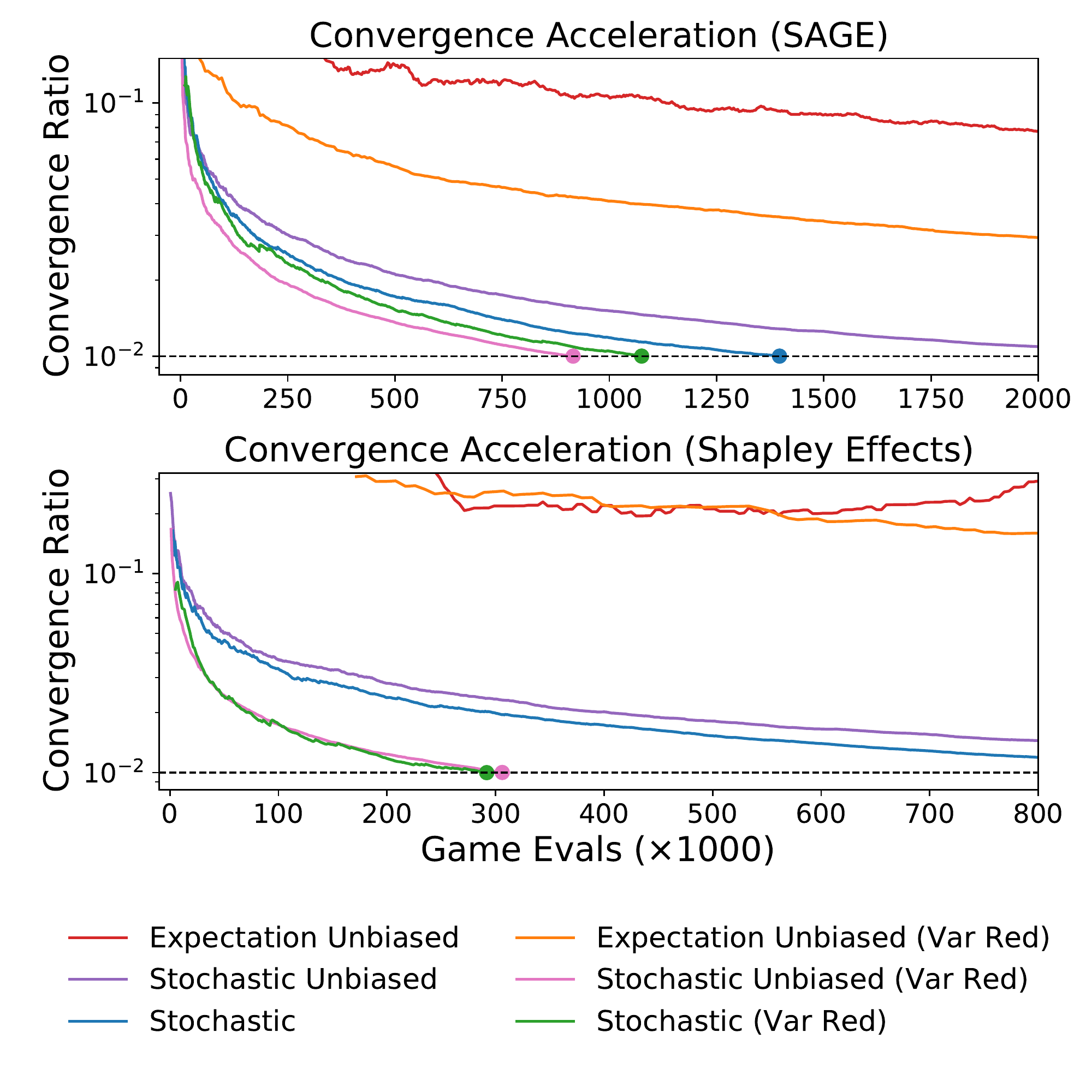}
\vskip -0.2in
\caption{Convergence acceleration for SAGE and Shapley Effects. The ratio of the maximum standard deviation to the gap between the largest and smallest Shapley values is compared across six estimators.}
\label{fig:acceleration}
\end{center}
\vskip -0.2in
\end{figure}

\begin{table*}
\caption{SHAP estimator run-time comparison. Each value represents the ratio of the average number of samples required relative to the fastest estimator for that dataset (lower is better).}
\label{tab:shap_comparison}
\vskip 0.1in
\begin{center}
\begin{small}
\begin{tabular}{lcccc}
\toprule
 & \textsc{Census Income} & \textsc{Bank Marketing} & \textsc{German Credit} & \textsc{BRCA Subtypes} \\
\midrule
Unbiased & 380.63 & 176.45 & 17437.44 & 90.40 \\
Unbiased + Paired Sampling & 128.60 & 90.61 & 422.17 & 40.44 \\
Original (KernelSHAP) & 12.74 & 7.41 & 13.74 & 2.49 \\
Original + Paired Sampling & \textbf{1.00} & \textbf{1.00} & \textbf{1.00} & \textbf{1.00} \\
\bottomrule
\end{tabular}
\end{small}
\end{center}
\end{table*}

To demonstrate local and global explanations with uncertainty estimates, we show examples of SHAP \cite{lundberg2017unified} and SAGE \cite{covert2020understanding} values generated using our estimators (Figure~\ref{fig:examples}). Both explanations used a convergence threshold of $t = 0.01$ and display 95\% confidence intervals, \textbf{which are features not previously offered by KernelSHAP}. We used the dataset sampling approach for both explanations, and for SAGE we used the estimator designed for stochastic cooperative games.
These estimators are faster than their unbiased versions, but the results are nearly identical.

To measure run-time differences between each estimator when calculating SHAP values, we compared the number of samples required to explain 100 instances for each dataset (Table~\ref{tab:shap_comparison}). Rather than reporting the exact number of samples, which is dependent on the convergence threshold, we show the \textit{ratio} between the number of samples required by each estimator; this ratio is independent of the convergence threshold when convergence is defined by the mean squared estimation error falling below a fixed value (Appendix~\ref{app:experiments}). Table~\ref{tab:shap_comparison} displays results based on 100 runs for each instance. Results show that the dataset sampling approach (original) is consistently faster than the unbiased estimator, and that paired sampling enables significantly faster convergence. In particular, we find that our paired sampling approach yields a \textbf{$\mathbf{9\times}$ speedup on average over the original KernelSHAP.}

To investigate the accuracy of our convergence forecasting method, we compared the predicted number of samples to the true number across 250 runs. The number of samples depends on the convergence threshold, and we used a threshold $t = 0.005$ for SHAP and $t = 0.02$ for Shapley Effects and SAGE. Figure~\ref{fig:forecasting} shows the results for SHAP (using the census data), Shapley Effects (using the bank data) and SAGE (using the BRCA data). In all three cases, \textbf{the forecasts become more accurate with more samples, and they vary within an increasingly narrow range around the true number of required samples.} There is a positive bias in the forecast, but the bias diminishes with more samples.

Finally, to demonstrate the speedup from our approach for stochastic cooperative games, we show that our stochastic estimator converges faster than a naive estimator based on the underlying game's expectation (see Section~\ref{sec:generalizing}). We plotted the ratio between the maximum standard deviation and the gap between the smallest and largest values, which we used to detect convergence (using a threshold $t = 0.01$).  Figure~\ref{fig:acceleration} shows that the stochastic approach dramatically speeds up both SAGE (using the BRCA data) and Shapley Effects (using the bank data), and that the paired sampling technique accelerates convergence for all estimators. The estimators based on the game's expectation are prohibitively slow and could not be run to convergence. \textbf{The fastest estimators for both datasets are stochastic estimators using the paired sampling technique,} and only these methods converged for both datasets in the number of samples displayed. As is the case for SHAP, the dataset sampling approach is often faster than the unbiased approach, but the latter is slightly faster for SAGE when using paired sampling.

\section{DISCUSSION}

This paper described several approaches for estimating Shapley values via linear regression. We first introduced an unbiased version of KernelSHAP, with  properties that are simpler to analyze than the original version. We then developed techniques for detecting convergence, calculating uncertainty estimates, and reducing the variance of both the original and unbiased estimators. Finally, we adapted our approach to provide significantly faster estimators for two global explanation methods based on stochastic cooperative games. Our work makes significant strides towards improving the practicality of Shapley value estimation by automatically determining the required number of samples, providing confidence intervals, and accelerating the estimation process.

More broadly, our work contributes to a mature literature on Shapley value estimation \cite{castro2009polynomial, maleki2013bounding} and to the growing ML model explanation field \cite{owen2014sobol, vstrumbelj2014explaining, datta2016algorithmic, lundberg2017unified, covert2020understanding}. We  focused on improving the regression-based approach to Shapley value estimation, and we leave to future work a detailed comparison of this approach to sampling-based \cite{vstrumbelj2014explaining, song2016shapley, chen2018shapley, covert2020understanding} and model-specific approximations \cite{ancona2019explaining, lundberg2020local}. We also believe that certain insights from our work may be applicable to LIME, which is based on a similar dataset sampling approach; recent work has noted LIME's high variance when using an insufficient number of samples \cite{bansal2020sam}, and an improved understanding of its convergence properties
\cite{garreau2020looking, mardaoui2020analysis} may lead to approaches for automatic convergence detection and uncertainty estimation.


\clearpage
\onecolumn
\appendix

\section[alternative title]{CALCULATING $A$ EXACTLY} \label{app:exact}

Recall the definition of $A$, which is a term in the solution to the Shapley value linear regression problem:

\begin{equation*}
    A = \E[ZZ^T].
\end{equation*}

The entries of $A$ are straightforward to calculate because $Z$ is a random binary vector with a known distribution. Recall that $Z$ is distributed according to $p(Z)$, which is defined as:

\begin{align*}
    p(z) = \begin{cases}
        Q\inv \mu_{\mathrm{Sh}}(Z) \quad 0 < \mathbf{1}^Tz < d \\
        0 \quad\quad\quad\quad\quad\; \mathrm{otherwise},
    \end{cases}
\end{align*}

where the normalizing constant $Q$ is given by:

\begin{align*}
    Q &= \sum_{0 < \mathbf{1}^Tz < d} \mu_{\mathrm{Sh}}(z) \\
    &= \sum_{k = 1}^{d - 1} \binom{d}{k} \frac{d - 1}{\binom{d}{k}k(d - k)} \\
    &= (d - 1) \sum_{k = 1}^{d - 1} \frac{1}{k(d - k)}.
\end{align*}

Although $Q$ does not have a simple closed-form solution, the expression above can be calculated numerically. The diagonal entries $A_{ii}$ are then given by:

\begin{align*}
    A_{ii} &= \E[Z_iZ_i]
    = p(Z_i = 1) \\
    &= \sum_{k = 1}^{d - 1} p(Z_i = 1 | \mathbf{1}^TZ = k) p(\mathbf{1}^TZ = k) \\
    &= \sum_{k = 1}^{d - 1} \frac{\binom{d - 1}{k - 1}}{\binom{d}{k}} \cdot Q\inv \binom{d}{k} \frac{d - 1}{\binom{d}{k}k(d - k)} \\
    &= \frac{\sum_{k = 1}^{d - 1} \frac{1}{d(d - k)}}{\sum_{k = 1}^{d - 1} \frac{1}{k(d - k)}}.
\end{align*}

This is equal to $\frac{1}{2}$ regardless of the value of $d$. To see this, consider the probability $p(Z_i = 0)$:

\begin{align*}
    p(Z_i = 0) &= 1 - p(Z_i = 1) \\
    &= 1 - \frac{\sum_{k = 1}^{d - 1} \frac{1}{d(d - k)}}{\sum_{k = 1}^{d - 1} \frac{1}{k(d - k)}} \\
    &= \frac{\sum_{k = 1}^{d - 1} \frac{1}{d(d - k)}}{\sum_{k = 1}^{d - 1} \frac{1}{k(d - k)}} \\
    &= p(Z_i = 1) \\
    \Rightarrow A_{ii} &= \frac{1}{2}.
\end{align*}

Next, consider the off-diagonal entries $A_{ij}$ for $i \neq j$:

\begin{align*}
    A_{ij} &= \E[Z_iZ_j]
    = p(Z_i = Z_j = 1) \\
    &= \sum_{k = 2}^{d - 1} p(Z_i = Z_j = 1 | \mathbf{1}^TZ = k) p(\mathbf{1}^TZ = k) \\
    &= \sum_{k = 2}^{d - 1} \frac{\binom{d - 2}{k - 2}}{\binom{d}{k}} \cdot Q\inv \binom{d}{k} \frac{d - 1}{\binom{d}{k}k(d - k)} \\
    &= \frac{1}{d(d - 1)}\frac{\sum_{k = 2}^{d - 1} \frac{k - 1}{d - k}}{\sum_{k = 1}^{d - 1} \frac{1}{k(d - k)}}.
\end{align*}

The value for off-diagonal entries $A_{ij}$ depends on $d$, unlike the diagonal entries $A_{ii}$. Although it does not have a simple closed-form expression, this value can be calculated numerically in $\mathcal{O}(d)$ time.

\section{VARIANCE REDUCTION PROOF} \label{app:variance_reduction}

We present a proof for Theorem~\ref{thm:confidence}, and we prove that a weaker condition than $G_v \succeq 0$ holds for all cooperative games (the diagonal elements satisfy $(G_v)_{ii} \geq 0$ for all games $v$).

\subsection{Theorem~\ref{thm:confidence} Proof}

In Section~\ref{sec:var_reduction}, we proposed a variance reduction technique that pairs each sample $z_i \sim p(Z)$ with its complement $\mathbf{1} - z_i$ when estimating $b$. We now provide a proof for the condition that must be satisfied for the estimator $\check \beta_n$ to have lower variance than $\bar \beta_n$. As mentioned in the main text, the multivariate CLT asserts that

\begin{align*}
    \bar b_n \sqrt{n} &\xrightarrow{D} \mathcal{N}(b, \Sigma_{\bar b}) \\
    \check b_n \sqrt{n} & \xrightarrow{D} \mathcal{N}(b, \Sigma_{\check b}),
\end{align*}

where

\begin{align*}
    \Sigma_{\bar b} &= \Cov\big(Z v(Z)\big), \\
    \Sigma_{\check b} &= \Cov\Big(\frac{1}{2} \big(Zv(Z) + (\mathbf{1} - Z)v(\mathbf{1} - Z) \big)\Big).
\end{align*}

We can also apply the multivariate CLT to the Shapley value estimators $\bar \beta_n$ and $\check \beta_n$. We can see that

\begin{align*}
    \bar \beta_n \sqrt{n} &\xrightarrow{D} \mathcal{N}(\beta^*, \Sigma_{\bar \beta}) \\
    \check \beta_n \sqrt{n} &\xrightarrow{D} \mathcal{N}(\beta^*, \Sigma_{\check \beta}),
\end{align*}

where, due to their multiplicative dependence on $b$ estimators, the covariance matrices are defined as

\begin{align*}
    \Sigma_{\bar \beta} &= C \Sigma_{\bar b} C^T \\
    \Sigma_{\check \beta} &= C \Sigma_{\check b} C^T.
\end{align*}

Next, we examine the relationship between $\Sigma_{\bar b}$ and $\Sigma_{\check b}$ because they dictate the relationship between $\Sigma_{\bar \beta}$ and $\Sigma_{\check \beta}$. To simplify our notation, we introduce three jointly distributed random variables, $M^0$, $M^1$ and $\bar M$, which are all functions of the random variable $Z$:

\begin{align*}
    M^0 &= Z v(Z) - \E[Z] v(\mathbf{0}) \\
    M^1 &= (\mathbf{1} - Z) v(\mathbf{1} - Z) - \E[\mathbf{1} - Z] v(\mathbf{0}) \\
    \bar M &= \frac{1}{2} \big( M^0 + M^1 \big).
\end{align*}

To understand $\bar M$'s covariance structure, we can decompose it using standard covariance properties and the fact that $p(z) = p(\mathbf{1} - z)$ for all $z$:

\begin{align*}
    \Cov(\bar M, \bar M)_{ij} &= \frac{1}{4} \Cov(M^0_i + M^1_i, M^0_j + M^1_j) \\
    &= \frac{1}{4} \Big( \Cov(M^0_i, M^0_j) + \Cov(M^1_i, \bar M^1_j) + \Cov(M^0_i, M^1_j) + \Cov(M^1_i, M^0_j) \Big) \\
    &= \frac{1}{2} \Big( \Cov(M^0_i, M^0_j) + \Cov(M^0_i, M^1_j) \Big).
\end{align*}

We can now compare $\Sigma_{\bar b}$ to $\Sigma_{\check b}$. To account for each $\bar M$ sample requiring twice as many cooperative game evaluations as $M^0$, we compare the covariance $\Cov(\bar b_{2n})$ to the covariance $\Cov(\check b_n)$:

\begin{align*}
    n \Big(\Cov(\bar b_{2n}) - \Cov(\check b_n) \Big)_{ij} &= - \frac{1}{2} \Cov(M^0_i, M^1_j).
\end{align*}

Based on this, we define $G_v$ as follows:

\begin{align*}
    G_v &= - \Cov(M^0_i, M^1_j) \\
    &= - \Cov\Big(Zv(Z) - \E[Z]v(\mathbf{0}), (\mathbf{1} - Z)v(\mathbf{1} - Z) - \E[\mathbf{1} - Z]v(\mathbf{0})\Big) \\
    &= - \Cov\Big(Zv(Z), (\mathbf{1} - Z)v(\mathbf{1} - Z)\Big).
\end{align*}

This is the matrix referenced in Theorem~\ref{thm:confidence}. Notice that $G_v$ is the negated cross-covariance between $M^0$ and $M^1$, which is the off-diagonal block in the joint covariance matrix for the concatenated random variable $(M^0, M^1)$. This matrix is symmetric, unlike general cross-covariance matrices, and its eigen-structure determines whether our variance reduction approach is effective. In particular, if the condition $G_v \succeq 0$ is satisfied, then we have

\begin{align*}
    \Cov(\bar b_{2n}) \succeq \Cov(\check b_n),
\end{align*}

which implies that

\begin{align*}
    \Cov(\bar \beta_{2n}) \succeq \Cov(\check \beta_n).
\end{align*}

Since the inverses of two ordered matrices are also ordered, we get the result:

\begin{align*}
    \Cov(\bar \beta_{2n})\inv \preceq \Cov(\check \beta_n)\inv.
\end{align*}

This has implications for quadratic forms involving each matrix. For any vector $a \in \R^d$, we have the inequality

\begin{align*}
    a^T\Cov(\bar \beta_{2n})\inv a \leq a^T\Cov(\check \beta_n)\inv a.
\end{align*}

The last inequality has a geometric interpretation. It shows that the confidence ellipsoid (i.e., the confidence region, or prediction ellipsoid) for $\check \beta_n$ is contained by the corresponding confidence ellipsoid for $\bar \beta_{2n}$ since large values of $n$ lead each estimator to converge to its asymptotically normal distribution. This is because the confidence ellipsoids are defined for $\alpha \in (0, 1)$ as

\begin{align*}
    & \bar E_{2n, \alpha} = \Big\{a \in \R^d: (a - \beta^*)^T \Cov(\bar \beta_{2n})\inv (a - \beta^*) \leq \sqrt{\chi^2_d(\alpha)} \Big\} \\
    & \check E_{n, \alpha} = \Big\{a \in \R^d: (a - \beta^*)^T \Cov(\check \beta_{n})\inv (a - \beta^*) \leq \sqrt{\chi^2_d(\alpha)} \Big\},
\end{align*}

where $\chi^2_d(\alpha)$ denotes the inverse CDF of a Chi-squared distribution with $d$ degrees of freedom evaluated at $\alpha$. More precisely, we have $\check E_{n, \alpha} \subseteq \bar E_{2n, \alpha}$ because

\begin{align*}
    &(a - \beta^*)^T \Cov(\check \beta_{n})\inv (a - \beta^*) \leq \sqrt{\chi^2_d(\alpha)} \\
    \Rightarrow &(a - \beta^*)^T \Cov(\bar \beta_{2n})\inv (a - \beta^*) \leq \sqrt{\chi^2_d(\alpha)}.
\end{align*}

This completes the proof.


\subsection{A Weaker Condition}

Consider the matrix $G_v$, which for a game $v$ is defined as

\begin{equation*}
    G_v = - \Cov\Big(Zv(Z), (\mathbf{1} - Z)v(\mathbf{1} - Z)\Big).
\end{equation*}

A necessary (but not sufficient) condition for $G_v \succeq 0$ is that its diagonal elements are non-negative. We can prove that this weaker condition holds for all games. For an arbitrary game $v$, the diagonal value $(G_v)_{ii}$ is given by:

\begin{align*}
    (G_v)_{ii} &= - \Cov\Big(Z_i v(Z), (1 - Z_i) v(\mathbf{1} - Z)\Big) \\
    &= - \E\big[Z_i(1 - Z_i)v(Z)v(\mathbf{1} - Z)\big] + \E\big[Z_iv(Z)\big]\E\big[(1 - Z_i)v(\mathbf{1} - Z)\big] \\
    &= \E\big[Z_iv(Z)\big]^2 \\
    &= \E\big[v(S) | i \in S\big]^2 \\
    &\geq 0.
\end{align*}

Geometrically, this condition means that the confidence ellipsoid $\bar E_{2n, \alpha}$ extends beyond the ellipsoid $\check E_{n, \alpha}$ in the axis-aligned directions. In a probabilistic sense, it means that the variance for each Shapley value estimate is lower when using the paired sampling technique.

\section{SHAPLEY EFFECTS} \label{app:shapley_effects}

Shapley Effects is a model explanation method that summarizes the model $f$'s sensitivity to each feature \cite{owen2014sobol}. It is based on the cooperative game

\begin{equation}
     \tilde w(S) = \Var\big(\E[f(X) | X_S]\big).
\end{equation}

To show that Shapley Effects can be viewed as the expectation of a stochastic cooperative game, we  reformulate this game (Covert et al. \cite{covert2020understanding}) as:

\begin{align*}
    \tilde w(S) &= \Var\big(\E[f(X) | X_S]\big) \\
    &= \Var\big(f(X)\big) - \E_{X_S}\big[\Var(f(X) | X_S)\big] \\
    &= c - \E_{X_S} \Big[ \E_{X_{D \setminus S} | X_S}\big[\big(\E[f(X) | X_S] - f(X_S, X_{D \setminus S}) \big)^2 \big] \Big] \\
    &= c - \E_X \Big[ \big( \E[f(X) | X_S] - f(X) \big)^2 \Big].
\end{align*}

If we generalize this cooperative game to allow arbitrary loss functions (e.g., cross entropy loss for classification tasks) rather than MSE, then we can ignore the constant value and re-write the game as

\begin{align*}
    \tilde w(S) = - \E_X\Big[ \ell\big(\E[f(X) | X_S], f(X)\big)\Big].
\end{align*}

Now, it is apparent that Shapley Effects is based on a cooperative game that is the expectation of a stochastic cooperative game, or $\tilde w(S) = \E_X[\tilde W(S, X)]$, where $\tilde W(S, X)$ is defined as:

\begin{equation*}
    \tilde W(S, X) = - \ell \big( \E[f(X) | X_S], f(X) \big).
\end{equation*}

Unlike the stochastic cooperative game implicitly used by SAGE, the exogenous random variable for this game is $U = X$.

\section{STOCHASTIC COOPERATIVE GAME PROOFS} \label{app:stochastic}


For a stochastic cooperative game $V(S, U)$, the generalized Shapley values are given by the expression

\begin{align*}
    \phi_i(V) &= \frac{1}{d} {\sum_{S {\subseteq} D {\setminus} \{i\}}} \binom{d {-} 1}{|S|}^{{-}1} \E_U\big[ V(S \cup \{i\}, U) {-} V(S, U) \big] \\
    &= \frac{1}{d} {\sum_{S {\subseteq} D {\setminus} \{i\}}} \binom{d {-} 1}{|S|}^{{-}1} \E_U\big[ V(S \cup \{i\}, U) \big] - \E_U\big[V(S, U) \big].
\end{align*}

The second line above shows that the generalized Shapley values are equivalent to the Shapley values of the game's expectation, or $\phi_i(\bar V)$, where $\bar V(S) = \E_U[V(S, U)]$. Based on this, we can also understand the values $\phi_1(V), \ldots, \phi_d(V)$ as the optimal coefficients for the following weighted least squares problem:

\begin{align*}
    &\min_{\beta_0, \ldots, \beta_d} \; \sum_z p(z) \Big(\beta_0 + z^T \beta - \E_U\big[V(z, U)\big] \Big)^2 \nonumber \\
    &\mathrm{s.t.} \quad \beta_0 = \E_U\big[V(\mathbf{0}, U) \big], \quad \mathbf{1}^T \beta = \E_U\big[V(\mathbf{1}, U) \big] - \E_U\big[V(\mathbf{0}, U) \big].
\end{align*}

Using our derivation from the main text (Section~\ref{sec:exact}),
we can write the solution as

\begin{equation*}
    \beta^* = A\inv \Big(b - \mathbf{1} \frac{\mathbf{1}^T A\inv b - \E_U[V(\mathbf{1}, U) ] + \E_U[V(\mathbf{0}, U) ]}{\mathbf{1}^T A\inv \mathbf{1}}\Big),
\end{equation*}

where $A$ and $b$ are given by the expressions

\begin{align*}
    A &= \E[ZZ^T] \\
    b &= \E_Z\Big[Z \big(\E_U[V(Z, U) ] - \E_U[V(\mathbf{0}, U) ]\big)\Big]. 
\end{align*}

Now, we consider our adaptations of KernelSHAP and unbiased KernelSHAP and examine whether these estimators are consistent or unbiased. We begin with the stochastic version of KernelSHAP presented in the main text (Section~\ref{sec:stochastic_kernelshap}). Recall that this approach uses the original $A$ estimator $\hat A_n$ and the modified $b$ estimator $\tilde b_n$, which is defined as:

\begin{equation*}
    \tilde b_n = \frac{1}{2} \sum_{i = 1}^n z_i \big( V(z_i, u_i) - \E_U\big[V(\mathbf{0}, U)\big] \big).
\end{equation*}

As mentioned in the main text, the strong law of large numbers lets us conclude that $\lim_{n \to \infty} \hat A_n = A$. Thus, we can understand the $b$ estimator's expectation as follows:

\begin{align*}
    \E\big[ \tilde b_n \big] &= \E_{ZU} \Big[ Z \big( V(Z, U) - \E_U\big[V(\mathbf{0}, U)\big] \big) \Big] \\
    &= \E_Z\Big[Z \big(\E_U[V(Z, U) ] - \E_U[V(\mathbf{0}, U) ]\big)\Big] \\
    &= b.
\end{align*}

With this, we conclude that $\lim_{n \to \infty} \tilde b_n = b$ and that $\tilde \beta_n$ are consistent, or

\begin{equation*}
    \lim_{n \to \infty} \tilde \beta_n = \beta^*.
\end{equation*}

To adapt unbiased KernelSHAP to the setting of stochastic cooperative games, we use the same technique of pairing independent samples of $Z$ and $U$.
To estimate $b$, we use an estimator $\tilde{\bar b}_n$ defined as:

\begin{equation*}
    \tilde{\bar b}_n = \frac{1}{n} \sum_{i = 1}^n z_iV(z_i, u_i) - \E\big[Z\big]\E_U\big[V(\mathbf{0}, U)\big].
\end{equation*}

We then substitute this into a Shapley value estimator as follows:

\begin{equation}
     \tilde{\bar \beta}_n = A\inv \Big(\tilde{\bar b}_n - \mathbf{1} \frac{\mathbf{1}^T A\inv \tilde{\bar b}_n - v(\mathbf{1}) + v(\mathbf{0})}{\mathbf{1}^T A\inv \mathbf{1}}\Big).
\end{equation}

This is consistent and unbiased because of the linear dependence on $\tilde{\bar b}_n$ and the fact that $\tilde{\bar b}_n$ is unbiased:

\begin{align*}
    \E\big[ \tilde{\bar b}_n \big] &= \E_{ZU} \Big[ ZV(Z, U) - \E\big[Z\big]\E_U\big[V(\mathbf{0}, U)\big] \Big] \\
    &= \E_Z\Big[Z \big(\E_U[V(Z, U) ] - \E_U[V(\mathbf{0}, U) ]\big)\Big] \\
    &= b.
\end{align*}

With this, we conclude that $\E[\tilde{\bar \beta}_n] = \beta^*$ and $\lim_{n \to \infty} \tilde{\bar \beta}_n = \beta^*$.

\section{EXPERIMENT DETAILS} \label{app:experiments}

Here, we provide further details about experiments described in the main body of text.

\subsection{Datasets and Hyperparameters}

For all three explanation methods considered in our experiments -- SHAP \cite{lundberg2017unified}, SAGE \cite{covert2020understanding} and Shapley Effects \cite{owen2014sobol} -- we handled removed features by marginalizing them out according to their joint marginal distribution. This is the default behavior for SHAP, but it is an approximation of what is required by SAGE and Shapley Effects. However, this choice should not affect the outcome of our experiments, which focus on the convergence properties of our Shapley value estimators (and not the underlying cooperative games).

Both SAGE and Shapley Effects require a loss function (Section~\ref{app:shapley_effects}). We used the cross entropy loss for SAGE and the soft cross entropy loss for Shapley Effects.

For the breast cancer (BRCA) subtype classification dataset, we selected 100 out of 17,814 genes to avoid overfitting on the relatively small dataset size (only 510 patients). These genes were selected at random: we tried ten random seeds and selected the subset that achieved the best performance to ensure that several relevant BRCA genes were included. A small portion of missing expression values were imputed with their mean. The data was centered and normalized prior to fitting a $\ell_1$ regularized logistic regression model; the regularization parameter was chosen using a validation set.



\subsection{SHAP Run-time Comparison}

To compare the run-time of various SHAP value estimators, we sought to compare the ratio of the mean number of samples required by each method. For a single example $x$ whose SHAP values are represented by $\beta^*$, the mean squared estimation error can be decomposed into the variance and bias as follows:

\begin{equation*}
    \E\big[||\hat \beta_n - \beta^*||^2\big] = \E\big[||\hat \beta_n - \E[\hat \beta_n]||^2\big] + \big|\big|\E[\hat \beta_n] - \beta^*\big|\big|^2.
\end{equation*}

Since we found that the error is dominated by variance rather than bias (Section~\ref{sec:properties}), we can make the following approximation to relate the error to the trace of the covariance matrix:

\begin{align}
    \E\big[||\hat \beta_n - \beta^*||^2\big] &= \E\big[||\hat \beta_n - \E[\hat \beta_n]||^2\big] + \big|\big|\E[\hat \beta_n] - \beta^*\big|\big|^2 \nonumber \\
    &\approx \E\big[||\hat \beta_n - \E[\hat \beta_n]||^2\big] \nonumber \\
    &= \Tr\Big( \Cov(\hat \beta_n) \Big). \label{eq:error_approximation}
\end{align}

If we define convergence based on the mean estimation error falling below a threshold value $t$, then the convergence condition is

\begin{equation*}
    \E\big[||\hat \beta_n - \beta^*||^2\big] \leq t.
\end{equation*}

Using our approximation (Eq.~\ref{eq:error_approximation}), we can see that this condition is approximately equivalent to

\begin{equation*}
    \E\big[||\hat \beta_n - \beta^*||^2\big] \approx \Tr\Big( \Cov(\hat \beta_n) \Big) \approx \frac{\Tr(\Sigma_{\hat \beta})}{n} \leq t.
\end{equation*}

For a given threshold $t$, the mean number of samples required to explain individual predictions is therefore based on the mean trace of the covariance matrix $\Sigma_{\hat \beta}$ (or the analogous covariance matrix for a different estimator). To compare two methods, we simply calculate the ratio of the mean trace of the covariance matrices. These ratios are reported in Table~\ref{tab:shap_comparison}, where each covariance matrix is calculated empirically across 100 runs with $n = 2048$ samples.

\section{CONVERGENCE EXPERIMENTS} \label{app:convergence_experiments}

In Section~\ref{sec:properties}, we empirically compared the bias and variance for the original and unbiased versions of KernelSHAP using a single census income prediction. The results (Figure~\ref{fig:error}) showed that both versions' estimation errors were dominated by variance rather than bias, and that the original version had significantly lower variance. To verify that this result is not an anomaly, we replicated it on multiple examples and across several datasets.

\begin{figure} 
\begin{center}
\includegraphics[width=\columnwidth]{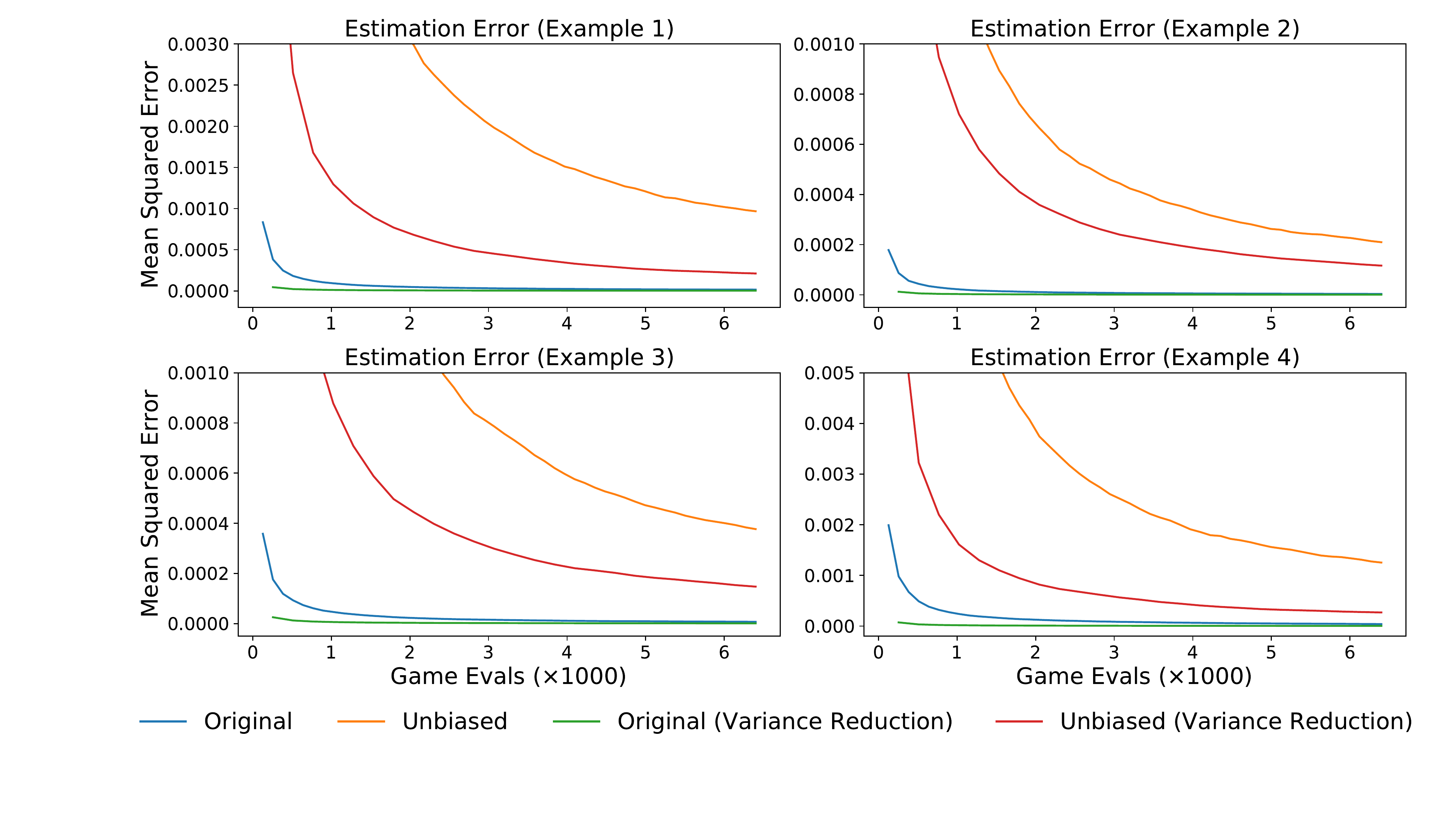}
\caption{Census income SHAP value estimation error on four predictions.}
\label{fig:census_mse}
\vskip 0.5in
\includegraphics[width=\columnwidth]{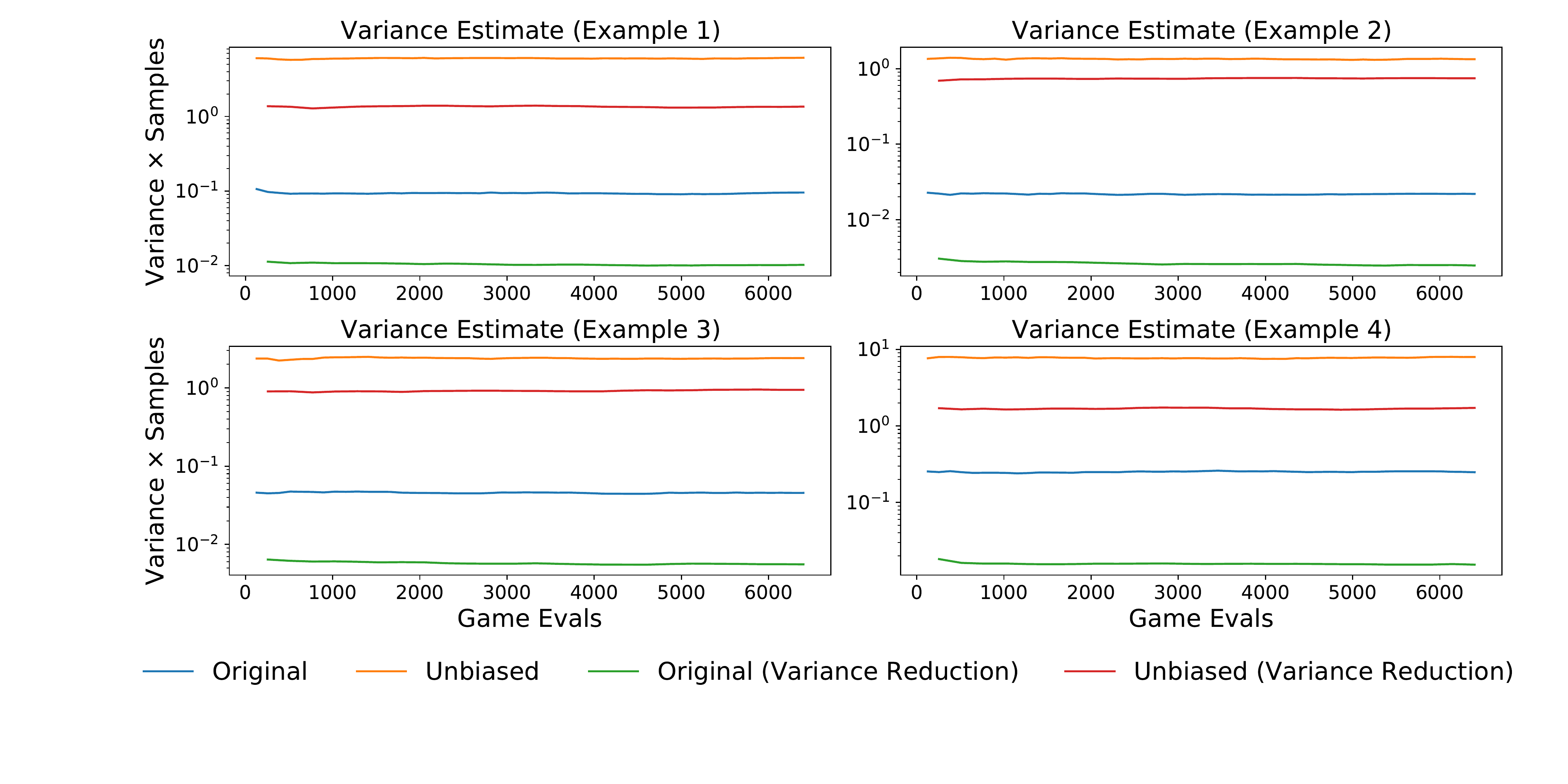}
\caption{Census income SHAP value variance estimation on four predictions.}
\label{fig:census_variance}
\end{center}
\end{figure}

\begin{figure} 
\begin{center}
\includegraphics[width=\columnwidth]{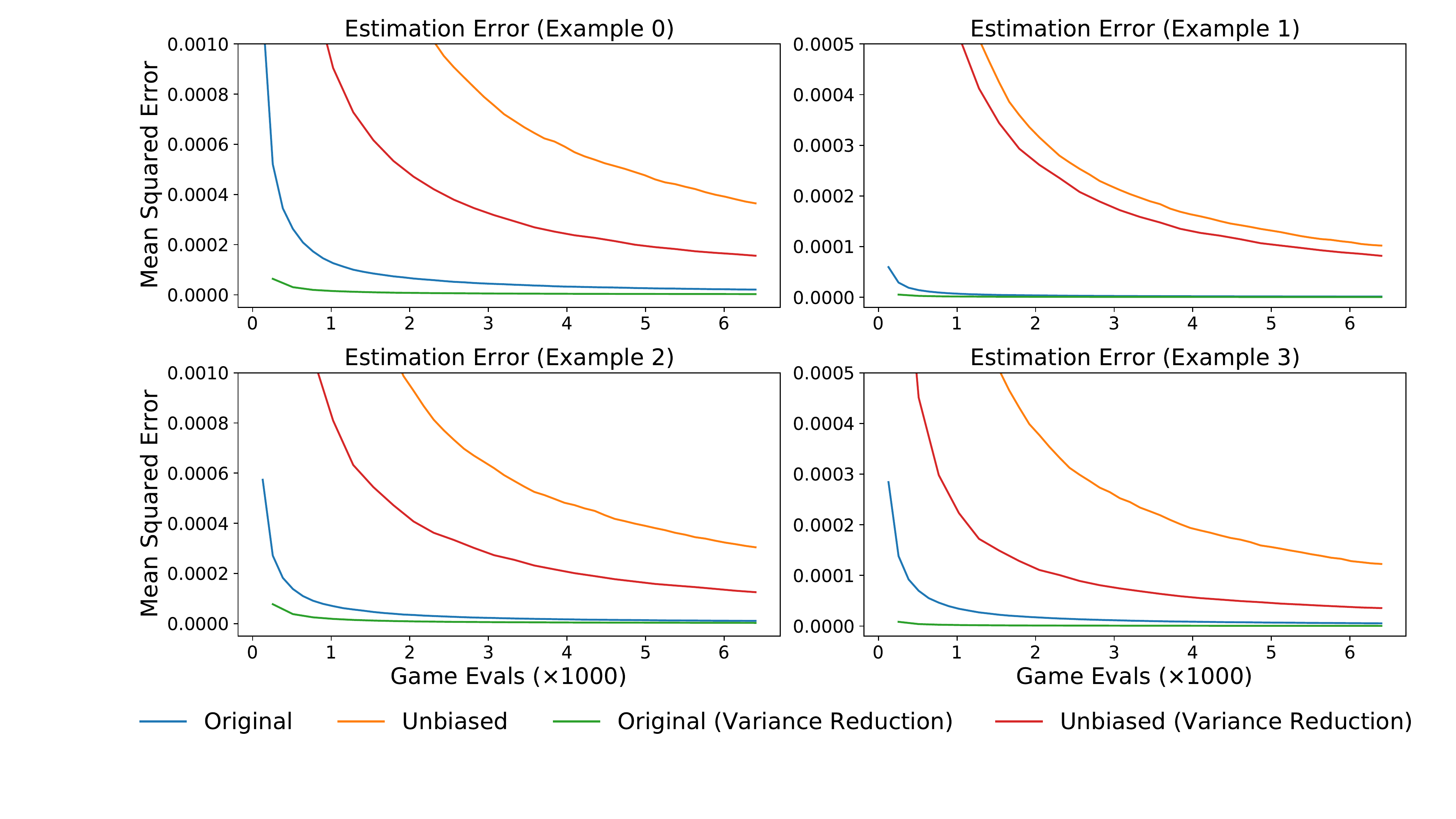}
\caption{Bank marketing SHAP value estimation error on four predictions.}
\label{fig:bank_mse}
\vskip 0.5in
\includegraphics[width=\columnwidth]{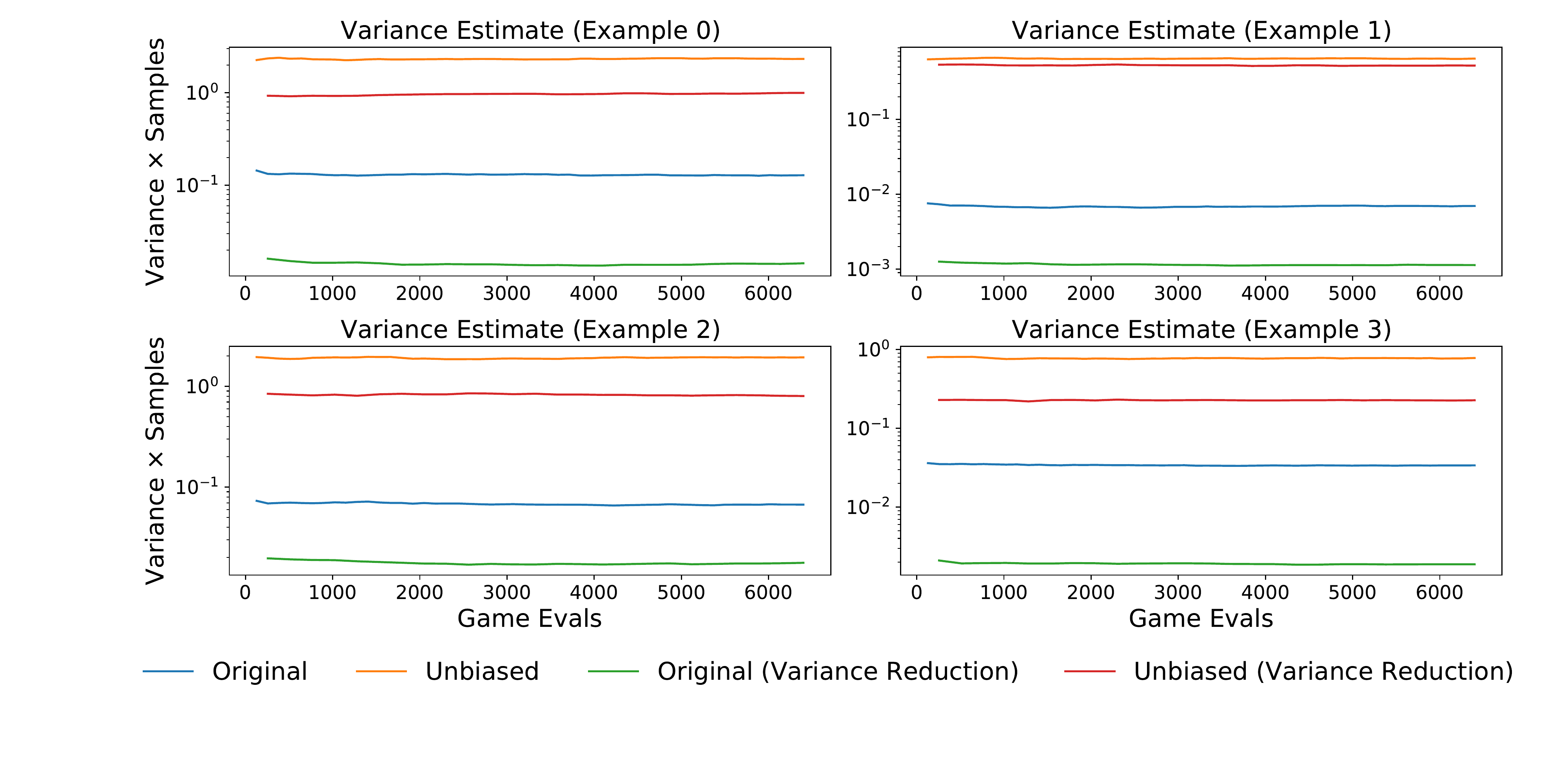}
\caption{Bank marketing SHAP value variance estimation on four predictions.}
\label{fig:bank_variance}
\end{center}
\end{figure}

\begin{figure} 
\begin{center}
\includegraphics[width=\columnwidth]{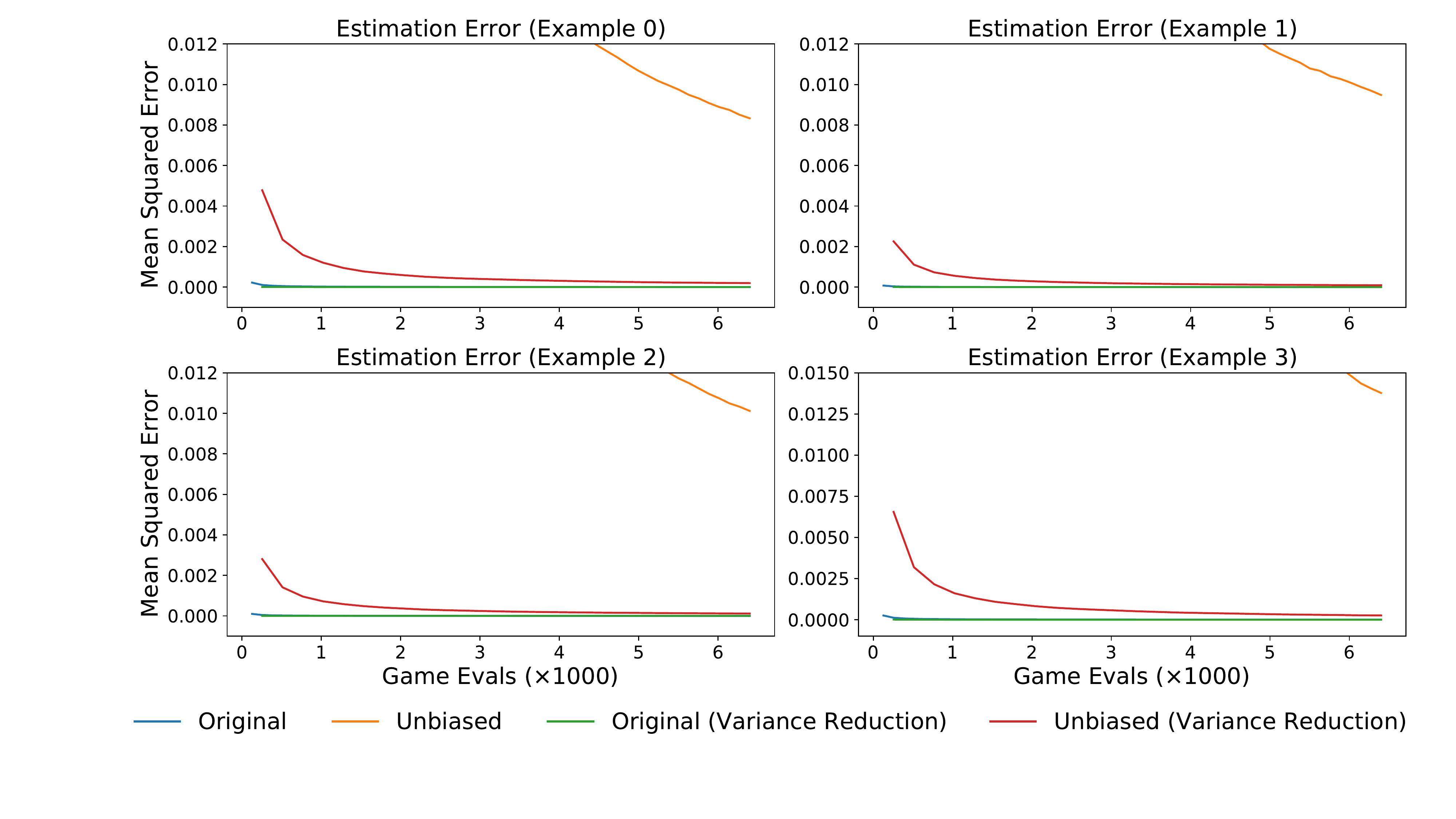}
\caption{German credit SHAP value estimation error on four predictions.}
\label{fig:credit_mse}
\vskip 0.5in
\includegraphics[width=\columnwidth]{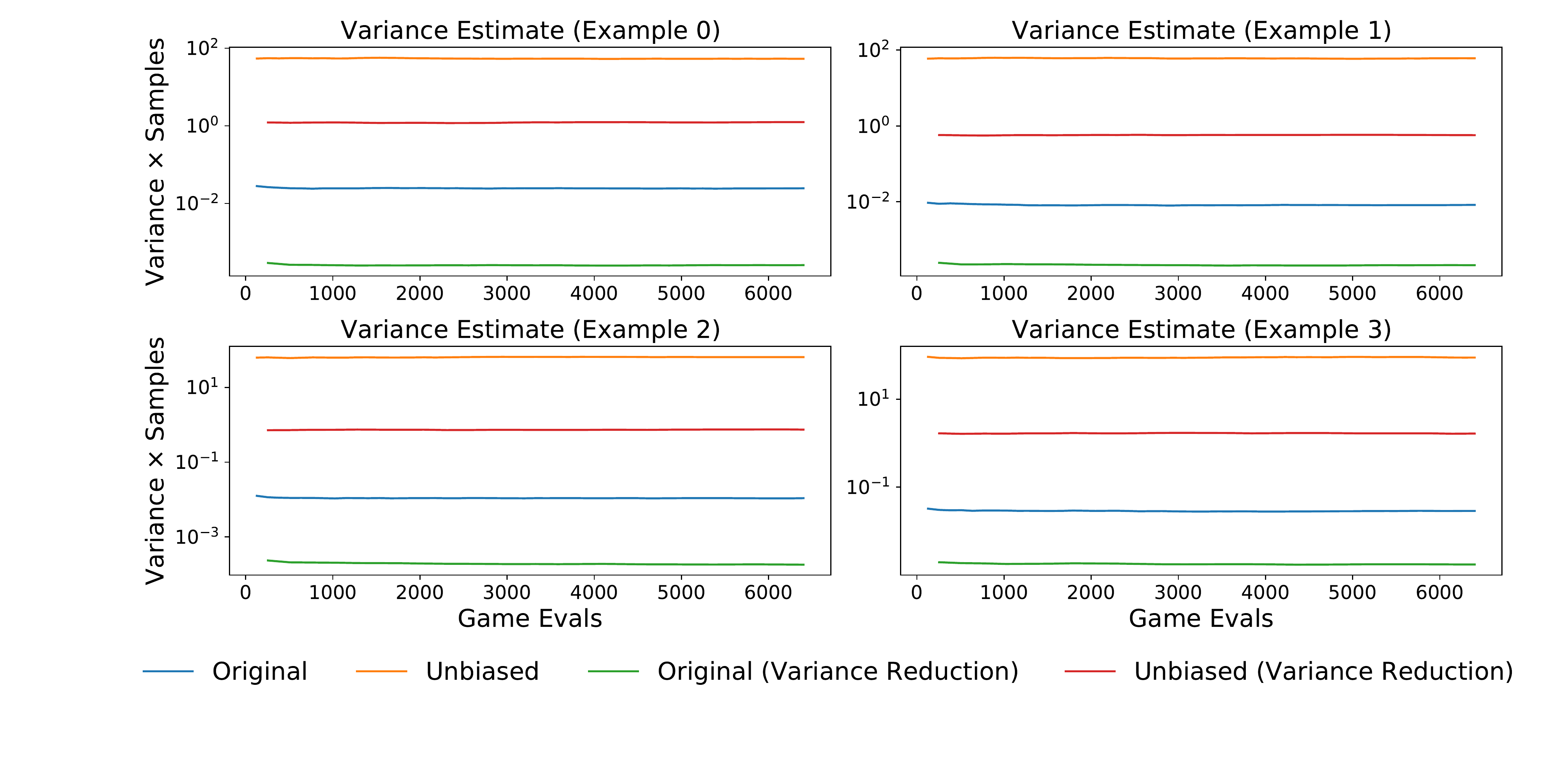}
\caption{German credit SHAP value variance estimation on four predictions.}
\label{fig:credit_variance}
\end{center}
\end{figure}

First, we examined several individual predictions for the census income, German credit and bank marketing datasets. To highlight the effectiveness of our paired sampling approach (Section~\ref{sec:var_reduction}), we added these methods as additional comparisons. Rather than decomposing the error into bias and variance as in the main text, we simply calculated the mean squared error across 100 runs of each estimator. Figure~\ref{fig:census_mse} shows the error for several census income predictions, Figure~\ref{fig:bank_mse} for several bank marketing predictions, and Figure~\ref{fig:credit_mse} for several credit quality predictions. These results confirm that the original version of KernelSHAP converges significantly faster than the unbiased version, and that the paired sampling technique is effective for both estimators. The dataset sampling approach (original KernelSHAP) appears preferable in practice despite being more difficult to analyze because it converges to the correct result much faster.

Second, we calculated a global measure of the bias and variance for each estimator using the same datasets (Table~\ref{tab:bias_variance}). Given 100 examples from each dataset, we calculated the mean bias and mean variance for each estimator empirically across 100 runs given $n = 256$
samples. Results show that the bias is nearly zero for all estimators, not just the unbiased ones; they also show that the variance is often significantly larger than the bias. However, when using the dataset sampling approach (original) in combination with the paired sampling technique, the bias and variance are comparably low ($\approx 0$) after 256
samples. The only exception is the unbiased estimator that does not use paired sampling, but this is likely due to estimation error because its bias is provably equal to zero.

Finally, Section~\ref{sec:convergence} also proposed assuming that the original KernelSHAP estimator's variance reduces at a rate of $\mathcal{O}(\frac{1}{n})$, similar to the unbiased version (for which we proved this rate). Although this result is difficult to prove formally, it seems to hold empirically across multiple predictions and several datasets. In Figures~\ref{fig:census_variance}, ~\ref{fig:bank_variance} and ~\ref{fig:credit_variance}, we display the product of the estimator's variance with the number of samples for the census, bank and credit datasets. Results confirm that the product is roughly constant as the number of samples increases, indicating that the variance for all four estimators (not just the unbiased ones) reduces at a rate of $\mathcal{O}(\frac{1}{n})$.


\begin{table*}
\caption{Global measures of bias and variance for each SHAP value estimator. Each entry is the mean bias and mean variance calculated empirically across 100 examples (bias/variance, lower is better).}
\label{tab:bias_variance}
\vskip 0.1in
\begin{center}
\begin{small}
\begin{tabular}{lccc}
\toprule
 & \textsc{Census Income} & \textsc{Bank Marketing} & \textsc{German Credit} \\
\midrule
Unbiased & 0.0002/0.0208 & 0.0001/0.0125 & 0.0026/0.2561 \\
Unbiased + Paired Sampling & 0.0000/0.0068 & 0.0000/0.0066 & 0.0000/0.0062 \\
Original (KernelSHAP) & 0.0000/0.0007 & 0.0000/0.0006 & 0.0000/0.0002 \\
Original + Paired Sampling & 0.0000/0.0001 & 0.0000/0.0001 & 0.0000/0.0000 \\
\bottomrule
\end{tabular}
\end{small}
\end{center}
\end{table*}


\section{ALGORITHMS} \label{app:algorithms}

Here, we provide pseudocode for the estimation algorithms described in the main text. Algorithm~\ref{alg:kernelshap} shows the dataset sampling approach (original KernelSHAP) with our convergence detection and paired sampling techniques. Algorithm~\ref{alg:kernelshap_stochastic} shows KernelSHAP's adaptation to the setting of stochastic cooperative games (stochastic KernelSHAP). Algorithm~\ref{alg:unbiased} shows the unbiased KernelSHAP estimator, and Algorithm~\ref{alg:unbiased_stochastic} shows the adaptation of unbiased KernelSHAP to stochastic cooperative games.

\begin{algorithm}[t]
\SetAlgoLined
\DontPrintSemicolon
\KwIn{Game $v$, convergence threshold $t$, intermediate samples $m$}
\tcp{Initialize}
n = 0\;
A = 0\;
b = 0\;
{}\;
\tcp{For tracking intermediate samples}
counter = 0\;
Atemp = 0\;
btemp = 0\;
estimates = list() \;
{}\;
\tcp{Sampling loop}
converged = False\;
\While{\upshape not converged}{
    \tcp{Draw next sample}
    Sample $z \sim p(Z)$\;
    \uIf{\upshape variance reduction}{
        Asample = $\frac{1}{2} \big( zz^T + (\mathbf{1} - z)(\mathbf{1} - z)^T \big)$\;
        bsample = $\frac{1}{2} \big(zv(z) + (\mathbf{1} {-} z)v(\mathbf{1} {-} z) - v(\mathbf{0})\big)$\;
    }
    \uElse{
        Asample = $zz^T$\;
        bsample = $z \big( v(z) - v(\mathbf{0})\big)$\;
    }
    {}\;
    \tcp{Welford's algorithm}
    n = n + 1\;
    A += (Asample $-$ A) / n \;
    b += (bsample $-$ b) / n\;
    counter += 1\;
    Atemp += (Asample $-$ Atemp) / counter \;
    btemp += (bsample $-$ btemp) / counter\;
    {}\;
    \uIf{\upshape counter == $m$}{
        \tcp{Get intermediate estimate}
        $\beta_m = \mathrm{Atemp}\inv \Big(\mathrm{btemp} - \mathbf{1} \frac{\mathbf{1}^T \mathrm{Atemp}\inv \mathrm{btemp} - v(\mathbf{1}) + v(\mathbf{0})}{\mathbf{1}^T \mathrm{Atemp}\inv \mathbf{1}}\Big)$\;
        estimates.append($\beta_m$)\;
        counter = 0\;
        Atemp = 0\;
        btemp = 0\;
        {}\;
        \tcp{Get estimates, uncertainties}
        $\beta_n = \mathrm{A}\inv \Big(\mathrm{b} - \mathbf{1} \frac{\mathbf{1}^T \mathrm{A}\inv \mathrm{b} - v(\mathbf{1}) + v(\mathbf{0})}{\mathbf{1}^T \mathrm{A}\inv \mathbf{1}}\Big)$\;
        $\Sigma_\beta = m \cdot \Cov(\mathrm{estimates})$ {   } \tcp{Empirical covariance}
        $\sigma_n = \sqrt{\mathrm{diag}(\Sigma_\beta) / \mathrm{n}}$ {   } \tcp{Element-wise square root}
        {}\;
        \tcp{Check for convergence}
        converged = $\Big( \frac{\max(\sigma_n)}{\max(\beta_n) - \min(\beta_n)} < t$ \Big)\;
    }
}
\KwRet{$\beta_n$, $\sigma_n$}
\caption{Shapley value estimation with dataset sampling (KernelSHAP)}
\label{alg:kernelshap}
\end{algorithm}

\begin{algorithm}[t]
\SetAlgoLined
\DontPrintSemicolon
\KwIn{Game $V$, convergence threshold $t$, intermediate samples $m$}
\tcp{Initialize}
n = 0\;
A = 0\;
b = 0\;
{}\;
\tcp{For tracking intermediate samples}
counter = 0\;
Atemp = 0\;
btemp = 0\;
estimates = list() \;
{}\;
\tcp{Sampling loop}
converged = False\;
\While{\upshape not converged}{
    \tcp{Draw next sample}
    Sample $z \sim p(Z)$\;
    Sample $u \sim p(U)$\;
    \uIf{\upshape variance reduction}{
        bsample = $\frac{1}{2} \big(zV(z, u) + (\mathbf{1} {-} z)V(\mathbf{1} {-} z, u) - \E_U[V(\mathbf{0}, U)]\big)$\;
        Asample = $\frac{1}{2} \big( zz^T + (\mathbf{1} - z)(\mathbf{1} - z)^T \big)$\;
    }
    \uElse{
        bsample = $z \big( V(z, u) - \E_U[V(\mathbf{0}, U)] \big)$\;
        Asample = $zz^T$\;
    }
    {}\;
    \tcp{Welford's algorithm}
    n = n + 1\;
    b += (bsample $-$ b) / n\;
    A += (Asample $-$ A) / n \;
    counter += 1\;
    btemp += (bsample $-$ btemp) / counter\;
    Atemp += (Asample $-$ Atemp) / counter \;
    {}\;
    \uIf{\upshape counter == $m$}{
        \tcp{Get intermediate estimate}
        $\beta_m = \mathrm{Atemp}\inv \Big(\mathrm{btemp} - \mathbf{1} \frac{\mathbf{1}^T \mathrm{Atemp}\inv \mathrm{btemp} - \E_U[V(\mathbf{1}, U)] + \E_U[V(\mathbf{0}, U)]}{\mathbf{1}^T \mathrm{Atemp}\inv \mathbf{1}}\Big)$\;
        estimates.append($\beta_m$)\;
        counter = 0\;
        Atemp = 0\;
        btemp = 0\;
        {}\;
        \tcp{Get estimates, uncertainties}
        $\beta_n = \mathrm{A}\inv \Big(\mathrm{b} - \mathbf{1} \frac{\mathbf{1}^T \mathrm{A}\inv \mathrm{b} - \E_U[V(\mathbf{1}, U)] + \E_U[V(\mathbf{0}, U)]}{\mathbf{1}^T \mathrm{A}\inv \mathbf{1}}\Big)$\;
        $\Sigma_\beta = m \cdot \Cov(\mathrm{estimates})$ {   } \tcp{Empirical covariance}
        $\sigma_n = \sqrt{\mathrm{diag}(\Sigma_\beta) / \mathrm{n}}$ {   } \tcp{Element-wise square root}
        {}\;
        \tcp{Check for convergence}
        converged = $\Big( \frac{\max(\sigma_n)}{\max(\beta_n) - \min(\beta_n)} < t$ \Big)\;
    }
}
\KwRet{$\beta_n$, $\sigma_n$}
\caption{Shapley value estimation with dataset sampling for stochastic cooperative games}
\label{alg:kernelshap_stochastic}
\end{algorithm}

\begin{algorithm}[t]
\SetAlgoLined
\DontPrintSemicolon
\KwIn{Game $v$, convergence threshold $t$}
\tcp{Initialize}
Set $A$ (Section~3.3)\;
Set $C$ (Eq.~13)\;
n = 0\;
b = 0\;
bSSQ = 0\;
{}\;
\tcp{Sampling loop}
converged = False\;
\While{\upshape not converged}{
    \tcp{Draw next sample}
    Sample $z \sim p(Z)$\;
    \uIf{\upshape variance reduction}{
        bsample = $\frac{1}{2} \big(zv(z) + (\mathbf{1} {-} z)v(\mathbf{1} {-} z) - v(\mathbf{0})\big)$\;
    }
    \uElse{
        bsample = $zv(z) - \frac{1}{2}v(\mathbf{0})$\;
    }
    {}\;
    \tcp{Welford's algorithm}
    n = n + 1\;
    diff = (bsample $-$ b) \;
    b += diff / n\;
    diff2 = (bsample $-$ b) \;
    bSSQ += outer(diff, diff2) {   } \tcp{Outer product}\;
    \tcp{Get estimates, uncertainties}
    $\beta_n = A\inv \Big(\mathrm{b} - \mathbf{1} \frac{\mathbf{1}^T A\inv \mathrm{b} - v(\mathbf{1}) + v(\mathbf{0})}{\mathbf{1}^T A\inv \mathbf{1}}\Big)$\;
    $\Sigma_{b} = $ bSSQ / n\;
    $\Sigma_{\beta} = C \Sigma_{b} C^T$\;
    $\sigma_n = \sqrt{\mathrm{diag}(\Sigma_{\beta}) / \mathrm{n}}$ {   } \tcp{Element-wise square root} \;
    \tcp{Check for convergence}
    converged = $\Big( \frac{\max(\sigma_n)}{\max(\beta_n) - \min(\beta_n)} < t$ \Big)\;
}
\KwRet{$\beta_n$, $\sigma_n$}
\caption{Unbiased Shapley value estimation}
\label{alg:unbiased}
\end{algorithm}

\begin{algorithm}[t]
\SetAlgoLined
\DontPrintSemicolon
\KwIn{Game $V$, convergence threshold $t$}
\tcp{Initialize}
Set $A$ (Section~3.3)\;
Set $C$ (Eq.~13)\;
n = 0\;
b = 0\;
bSSQ = 0\;
{}\;
\tcp{Sampling loop}
converged = False\;
\While{\upshape not converged}{
    \tcp{Draw next sample}
    Sample $z \sim p(Z)$\;
    Sample $u \sim p(U)$\;
    \uIf{\upshape variance reduction}{
        bsample = $\frac{1}{2} \Big(zV(z, u) + (\mathbf{1} {-} z)V(\mathbf{1} {-} z, u) - \E_U\big[ V(\mathbf{0}), U \big] \Big)$\;
    }
    \uElse{
        bsample = $zV(z, u) - \frac{1}{2} \E_U[V(\mathbf{0}, U)]$\;
    }
    {}\;
    \tcp{Welford's algorithm}
    n = n + 1\;
    diff = (bsample $-$ b) \;
    b += diff / n\;
    diff2 = (bsample $-$ b) \;
    bSSQ += outer(diff, diff2) {   } \tcp{Outer product}\;
    \tcp{Get estimates, uncertainties}
    $\beta_n = A\inv \Big(\mathrm{b} - \mathbf{1} \frac{\mathbf{1}^T A\inv \mathrm{b} - \E_U[V(\mathbf{1}, U)] + \E_U[V(\mathbf{0}, U)]}{\mathbf{1}^T A\inv \mathbf{1}}\Big)$\;
    $\Sigma_{b} = $ bSSQ / n\;
    $\Sigma_{\beta} = C \Sigma_{b} C^T$\;
    $\sigma_n = \sqrt{\mathrm{diag}(\Sigma_{\beta}) / \mathrm{n}}$ {   } \tcp{Element-wise square root} \;
    \tcp{Check for convergence}
    converged = $\Big( \frac{\max(\sigma_n)}{\max(\beta_n) - \min(\beta_n)} < t$ \Big)\;
}
\KwRet{$\beta_n$, $\sigma_n$}
\caption{Unbiased Shapley value estimation for stochastic cooperative games}
\label{alg:unbiased_stochastic}
\end{algorithm}

\clearpage

\subsubsection*{Acknowledgements}
This work was funded by the National Science Foundation [CAREER DBI-1552309, and DBI- 1759487]; the American Cancer Society [127332-RSG-15-097-01-TBG]; and the National Institutes of Health [R35 GM 128638, and R01 NIA AG 061132]. We would like to thank Hugh Chen, the Lee Lab, and our AISTATS reviewers for feedback that greatly improved this work.

\bibliographystyle{plainnat}
\bibliography{aistats_main}

\vfill

\end{document}


%
\runningtitle{Shapley Value Estimation Using Linear Regression}

%
\runningauthor{Covert \& Lee}

\onecolumn
\aistatstitle{Fixing KernelSHAP: Practical Shapley Value Estimation \\ Using Linear Regression
(Supplementary Materials)}


\section[alternative title]{CALCULATING $A$ EXACTLY}

Recall the definition of $A$, which acts as a covariance matrix in the Shapley value linear regression problem:

\begin{equation*}
    A = \E[ZZ^T].
\end{equation*}

The entries of $A$ are straightforward to calculate because $Z$ is a random binary vector. Recall that $Z$ is distributed according to $p(Z)$, which is defined as:

\begin{align*}
    p(z) = \begin{cases}
        \alpha\inv \mu_{\mathrm{Sh}}(Z) \quad 0 < \mathbf{1}^Tz < d \\
        0 \quad\quad\quad\quad\quad\; \mathrm{else}
    \end{cases}
\end{align*}

where the normalization constant $\alpha$ is given by:

\begin{align*}
    \alpha &= \sum_{0 < \mathbf{1}^Tz < d} \mu_{\mathrm{Sh}}(z) \\
    &= \sum_{k = 1}^{d - 1} \binom{d}{k} \frac{d - 1}{\binom{d}{k}k(d - k)} \\
    &= (d - 1) \sum_{k = 1}^{d - 1} \frac{1}{k(d - k)}.
\end{align*}

Although $\alpha$ does not have a simple closed-form solution, the expression above can be calculated numerically. We now return to $A$. The diagonal entries $A_{ii}$ are given by:

\begin{align*}
    A_{ii} &= \E[Z_iZ_i]
    = p(Z_i = 1) \\
    &= \sum_{k = 1}^{d - 1} p(Z_i = 1 | \mathbf{1}^TZ = k) p(\mathbf{1}^TZ = k) \\
    &= \sum_{k = 1}^{d - 1} \frac{\binom{d - 1}{k - 1}}{\binom{d}{k}} \cdot \alpha\inv \binom{d}{k} \frac{d - 1}{\binom{d}{k}k(d - k)} \\
    &= \frac{\sum_{k = 1}^{d - 1} \frac{1}{d(d - k)}}{\sum_{k = 1}^{d - 1} \frac{1}{k(d - k)}}
\end{align*}

It turns out that this is equal to one half, regardless of the value of $d$. To show this, consider how we would calculate the probability $p(Z_i = 0)$:

\begin{align*}
    p(Z_i = 0) &= 1 - p(Z_i = 1) \\
    &= 1 - \frac{\sum_{k = 1}^{d - 1} \frac{1}{d(d - k)}}{\sum_{k = 1}^{d - 1} \frac{1}{k(d - k)}} \\
    &= \frac{\sum_{k = 1}^{d - 1} \frac{1}{d(d - k)}}{\sum_{k = 1}^{d - 1} \frac{1}{k(d - k)}} \\
    &= p(Z_i = 1) \\
    \Rightarrow A_{ii} &= \frac{1}{2}
\end{align*}

Next, consider the off-diagonal entries $A_{ij}$ for $i \neq j$:

\begin{align*}
    A_{ij} &= \E[Z_iZ_j]
    = p(Z_i = Z_j = 1) \\
    &= \sum_{k = 2}^{d - 1} p(Z_i = Z_j = 1 | \mathbf{1}^TZ = k) p(\mathbf{1}^TZ = k) \\
    &= \sum_{k = 2}^{d - 1} \frac{\binom{d - 2}{k - 2}}{\binom{d}{k}} \cdot \alpha\inv \binom{d}{k} \frac{d - 1}{\binom{d}{k}k(d - k)} \\
    &= \frac{1}{d(d - 1)}\frac{\sum_{k = 2}^{d - 1} \frac{k - 1}{d - k}}{\sum_{k = 1}^{d - 1} \frac{1}{k(d - k)}}
\end{align*}

The value for off-diagonal entries $A_{ij}$ depends on $d$, and although it does not have a simple closed-form expression, it can also be calculated numerically.

\section{VARIANCE REDUCTION PROOFS}

In the main text, we present a variance reduction technique that pairs each sample $z_i \sim p(Z)$ with its complement $\mathbf{1} - z_i$ when estimating $b$. We now prove the result about the estimator $\tilde \beta_n$ having lower variance than $\bar \beta_n$.
As mentioned in the main text, the multivariate CLT asserts that

\begin{align*}
    b_n \sqrt{n} &\xrightarrow{D} \mathcal{N}(b, \Sigma_b), \\
    \tilde b_n \sqrt{n} & \xrightarrow{D} \mathcal{N}(b, \Sigma_{\tilde b})
\end{align*}

where

\begin{align*}
    \Sigma_b &= \Cov\Big(Z \big(v(Z) - v(\mathbf{0})\big)\Big) \\
    \Sigma_{\tilde b} &= \Cov\Big(\frac{1}{2} \big(Zv(Z) + (\mathbf{1} - Z)v(\mathbf{1} - Z) - v(\mathbf{0}) \big)\Big).
\end{align*}

Due to their multiplicative dependence on $b$ estimators, we can also apply the multivariate CLT to $\bar \beta_n$ and $\tilde \beta_n$:

\begin{align*}
    \bar \beta_n \sqrt{n} &\xrightarrow{D} \mathcal{N}(\beta^*, \Sigma_\beta) \\
    \tilde \beta_n \sqrt{n} &\xrightarrow{D} \mathcal{N}(\beta^*, \Sigma_{\tilde \beta})
\end{align*}

where we have

\begin{align*}
    \Sigma_\beta &= C \Sigma_b C^T \\
    \Sigma_{\tilde \beta} &= C \Sigma_{\tilde b} C^T.
\end{align*}

We proceed by decomposing the entries of $\Sigma_b$ and $\Sigma_{\tilde b}$ and build towards the eventual comparison of $\Sigma_\beta$ and $\Sigma_{\tilde \beta}$. To simplify our notation, we introduce the random variables $M$ and $\tilde M$:

\begin{align*}
    M &= Z\big(v(Z) - v(\mathbf{0})\big) \\
    \tilde M &= \frac{1}{2} \Big(Zv(Z) + (\mathbf{1} {-} Z) v(\mathbf{1} {-} Z) + v(\mathbf{0}) \Big).
\end{align*}





We can gain more insight into the entries of $\Sigma_b$ and $\Sigma_{\tilde b}$ by decomposing them using the law of total variance (or covariance). To do so, we condition on either $Z_i$ or the pair of variables $(Z_i, Z_j)$. To start, consider the diagonal entries $(\Sigma_b)_{ii}$:

\begin{align*}
    (\Sigma_b)_{ii} &= \Var(M_i) \\
    &= \E\big[\Var(M_i | Z_i)\big] + \Var\big(\E[M_i | Z_i]\big) \\
    &= \frac{1}{2} \Var\big(v(Z) | Z_i = 1\big) + \frac{1}{4} \E\big[v(Z) - v(\mathbf{0}) | Z_i = 1\big]^2.
\end{align*}

For off-diagonal entries (where $i \neq j$), we have: 

\begin{align*}
    (\Sigma_b)_{ij} &= \Cov(M_i, M_j) \\
    &= \E\big[\Cov(M_i, M_j | Z_iZ_j)\big] + \Cov\big(\E[M_i | Z_iZ_j], \E[M_j | Z_iZ_j]\big) \\
    &= p(Z_i = Z_j = 1) \Var\big(v(Z) | Z_i = Z_j = 1\big) \\
    &\quad + p(Z_i = Z_j = 1) \E[v(Z) - v(\mathbf{0}) | Z_i = Z_j = 1]^2
    - \frac{1}{4} \E\big[v(Z) - v(\mathbf{0}) | Z_i = 1\big]
    \E\big[v(Z) - v(\mathbf{0}) | Z_j = 1\big].
\end{align*}

Several of the terms in the expressions above are properties of the game, e.g., the variance of the game's value when a given player is included. Not all of these terms need to be known, because we are focused primarily on a comparison between $\Sigma_b$ and $\Sigma_{\tilde b}$. We now analyze $\Sigma_{\tilde b}$, finding that several of the same terms appear. For the diagonal entries, we can write:

\begin{align*}
    (\Sigma_{\tilde b})_{ii} &= \Var(\tilde M_i) \\
    &= \Var(\frac{1}{2} M_i | Z_i = 1) \\
    &= \frac{1}{4} \Var(v(Z) | Z_i = 1). \\
\end{align*}

To write expressions for the off-diagonal entries using the law of total covariance, we must resolve two complicated terms. We consider each term individually:

\begin{align*}
    \E\big[\Cov(\tilde M_i, \tilde M_j | Z_iZ_j)\big] &= \sum_{Z_iZ_j} p(Z_iZ_j) \Cov(\tilde M_i, \tilde M_j | Z_iZ_j) \\
    &= \frac{1}{2} p(Z_i = Z_j = 1) \Var(v(Z) | Z_i = Z_j = 1) \\
    &\quad + \frac{1}{2} p(Z_i = 1, Z_j = 0) \Cov\big(v(Z), v(\mathbf{1} - Z) | Z_i = 1, Z_j = 0\big) \\
    &= \frac{1}{2} p(Z_i = Z_j = 1) \Var(v(Z) | Z_i = Z_j = 1) \\
    %
    &\quad + \frac{1}{2} p(Z_i = 1, Z_j = 0) \E\Big[\big(v(Z) - v(\mathbf{0})\big)\big(v(\mathbf{1} - Z) - v(\mathbf{0})\big) | Z_i = 1, Z_j = 0\Big] \\
    &\quad - \frac{1}{2} p(Z_i = 1, Z_j = 0) \E\big[v(Z) - v(\mathbf{0}) | Z_i = 1, Z_j = 0\big] \E\big[v(Z) - v(\mathbf{0}) | Z_i = 0, Z_j = 1\big]
\end{align*}

\begin{align*}
    \Cov\big(\E[\tilde M_i | Z_iZ_j], \E[\tilde M_j | Z_iZ_j]\big) &= \E\Big[ \E[\tilde M_i | Z_iZ_j] \E[\tilde M_j | Z_iZ_j] \Big] - \E[\tilde M_i] \E[\tilde M_j] \\
    &= \frac{1}{2} p(Z_i = Z_j = 1) \E[v(Z) - v(\mathbf{0}) | Z_i = Z_j = 1]^2 \\
    %
    %
    &\quad + \frac{1}{2} p(Z_i = 1, Z_j = 0) \E\big[v(Z) - v(\mathbf{0}) | Z_i = 1, Z_j = 0\big] \E\big[v(Z) - v(\mathbf{0}) | Z_i = 0, Z_j = 1\big] \\
    &\quad - \frac{1}{4}\E\big[v(Z) - v(\mathbf{0}) | Z_i = 1\big]\E\big[v(Z) - v(\mathbf{0}) | Z_j = 1\big]
\end{align*}

Putting these together, the law of total covariance gives us:

\begin{align*}
    (\Sigma_{\tilde \beta})_{ij} &= \Cov(\tilde M_i, \tilde M_j) \\
    &= \E\big[\Cov(\tilde M_i, \tilde M_j | Z_iZ_j)\big] + \Cov\big(\E[\tilde M_i | Z_iZ_j], \E[\tilde M_j | Z_iZ_j]\big) \\
    &= \frac{1}{2} p(Z_i = Z_j = 1) \Var(v(Z) | Z_i = Z_j = 1) \\
    &\quad + \frac{1}{2} p(Z_i = 1, Z_j = 0) \E\Big[\big(v(Z) - v(\mathbf{0})\big)\big(v(\mathbf{1} - Z) - v(\mathbf{0})\big) | Z_i = 1, Z_j = 0\Big] \\
    &\quad + \frac{1}{2} p(Z_i = Z_j = 1) \E[v(Z) - v(\mathbf{0}) | Z_i = Z_j = 1]^2 \\
    &\quad - \frac{1}{4}\E\big[v(Z) - v(\mathbf{0}) | Z_i = 1\big]\E\big[v(Z) - v(\mathbf{0}) | Z_j = 1\big]
\end{align*}


With this, we can now compare $\Sigma_b$ and $\Sigma_{\tilde b}$. To account for the fact that the variance reduction technique requires twice as many cooperative game evaluations per sample (which is generally the most expensive part of this process), we compare the variance of the $b$ estimators $b_{2n}$ and $\tilde b_n$. We can consider the diagonal and off-diagonal element simultaneously because of similarities in their expressions. By allowing certain terms to cancel out, we get the following result:




\begin{align*}
    n \Big( \Cov(b_{2n}) - \Cov(\tilde b_n) \Big)_{ij} &= \Big(\frac{1}{2} \Sigma_b - \Sigma_{\tilde b} \Big)_{ij} \\
    &= \frac{1}{8} \E\big[v(Z) - v(\mathbf{0}) | Z_i = 1\big]\E\big[v(Z) - v(\mathbf{0}) | Z_j = 1\big] \\
    &\quad - \frac{1}{2} p(Z_i = 1, Z_j = 0) \E\Big[\big(v(Z) - v(\mathbf{0})\big)\big(v(\mathbf{1} - Z) - v(\mathbf{0})\big) | Z_i = 1, Z_j = 0\Big] \\
    &= \frac{1}{2} p(Z_i = 1) \E\big[v(Z) - v(\mathbf{0}) | Z_i = 1\big] p(Z_j = 1)\E\big[v(Z) - v(\mathbf{0}) | Z_j = 1\big] \\
    &\quad - \frac{1}{2} p(Z_i = 1, Z_j = 0) \E\Big[\big(v(Z) - v(\mathbf{0})\big)\big(v(\mathbf{1} - Z) - v(\mathbf{0})\big) | Z_i = 1, Z_j = 0\Big] \\
    %
    &= - \frac{1}{2} \Cov\Big(Z_i \big(v(Z) - v(\mathbf{0}) \big), \big(\mathbf{1} - Z_j \big) \big(v(\mathbf{1} - Z) - v(\mathbf{0}) \big) \Big)
\end{align*}

\textcolor{purple}{This gets weirder and weirder. It's indeed a covariance, but I can't exactly write this as a big covariance matrix. And it's negative covariance... Meaning that I better hope these things have negative correlation.}

\textcolor{purple}{Consider a random variable that concatenates $M$ with its complement vector $(1 - Z)(v(1 - Z) - v(0))$, and the random variable $\tilde M$ that is their mean. It's sort of like a control variate approach. What we are looking at, in the difference of their covariances, is a negated off-diagonal block from the joint covariance matrix.}

\textcolor{purple}{A good first step would be to show this in a much faster way.}

Putting these results together, we can see that the difference between the two covariance matrices is positive-semi definite. For any vector $a \in \R^d$, we have:

\begin{align*}
    a^T \Big( n\Cov(b_{2n}) - n\Cov(\tilde b_n) \Big) a &= \frac{1}{2} \Big( \sum_{i = 1}^d a_i p(Z_i = 1) \E\big[v(Z) - v(\mathbf{0}) | Z_i = 1\big] \Big)^2 \\
    &\quad - \frac{1}{2} \sum_{ij} a_i a_j p(Z_i = 1, Z_j = 0) \E\Big[\big(v(Z) - v(\mathbf{0})\big)\big(v(\mathbf{1} - Z) - v(\mathbf{0})\big) | Z_i = 1, Z_j = 0\Big] \\
    &= \frac{1}{2} \sum_{ij} a_i a_j p(Z_i = 1) p(Z_j = 1) \E\big[v(Z) - v(\mathbf{0}) | Z_i = 1\big] \E\big[v(Z) - v(\mathbf{0}) | Z_j = 1\big] \\
    &\quad - \frac{1}{2} \sum_{ij} a_i a_j p(Z_i = 1, Z_j = 0) \E\Big[\big(v(Z) - v(\mathbf{0})\big)\big(v(\mathbf{1} - Z) - v(\mathbf{0})\big) | Z_i = 1, Z_j = 0\Big] \\
    &= \frac{1}{2} \sum_{ij} a_i a_j \Big( p(Z_i = 1) p(Z_j = 1) \E\big[v(Z) - v(\mathbf{0}) | Z_i = 1\big] \E\big[v(Z) - v(\mathbf{0}) | Z_j = 1\big] \\
    &\quad\quad\quad\quad\quad\quad - p(Z_i = 1, Z_j = 0) \E\Big[\big(v(Z) - v(\mathbf{0})\big)\big(v(\mathbf{1} - Z) - v(\mathbf{0})\big) | Z_i = 1, Z_j = 0\Big] \Big) \\
    &= 
\end{align*}

This allows us to conclude that $\Cov(b_{2n}) \succeq \Cov(\tilde b_n)$, and therefore that $\Cov(\bar \beta_{2n}) \succeq \Cov(\tilde \beta_n)$. This completes the proof.

\vfill